\documentclass{article}

\usepackage{authblk}
\usepackage{arxiv}

\usepackage[utf8]{inputenc} 
\usepackage[T1]{fontenc}    
\usepackage{url}            
\usepackage{booktabs}       
\usepackage{amsfonts}       
\usepackage{nicefrac}       
\usepackage{microtype}      
\usepackage{lipsum}		
\usepackage{graphicx}
\usepackage[square,sort,comma,numbers]{natbib}
\usepackage{doi}

\usepackage{adjustbox}
\usepackage{amsmath}
\usepackage{wrapfig}

\usepackage[utf8]{inputenc} 
\usepackage[T1]{fontenc}    
\usepackage{url}            
\usepackage{booktabs}       
\usepackage{amsfonts}       
\usepackage{nicefrac}       
\usepackage{microtype}      
\usepackage[table]{xcolor}

\usepackage{lipsum}		
\usepackage{graphicx}
\usepackage{kotex}
\usepackage{bbding}
\usepackage{multirow}

\usepackage{array}
\usepackage{tabulary}
\usepackage{ctable}
\usepackage{subcaption}
\usepackage{amssymb}

\makeatletter
\newcommand{\thickhline}{%
    \noalign {\ifnum 0=`}\fi \hrule height 1pt
    \futurelet \reserved@a \@xhline
}

\newcolumntype{"}{@{\hskip\tabcolsep\vrule width 1pt\hskip\tabcolsep}}
\makeatother

\definecolor{Gray}{gray}{0.9}
\definecolor{LightCyan}{rgb}{0.88,1,1}

\usepackage[noend]{algpseudocode} 
\usepackage[ruled,vlined]{algorithm2e}

\definecolor{commentcolor}{RGB}{110,154,155}   
\newcommand{\PyComment}[1]{\ttfamily\textcolor{commentcolor}{\# #1}}  
\newcommand{\PyCode}[1]{\ttfamily\textcolor{black}{#1}} 

\title{Re-Scoring Using Image-Language Similarity for Few-Shot Object Detection}

\author[]{Min Jae Jung}
\author[]{Seung Dae Han}
\author[]{Joohee Kim}
\affil[]{AI Lab, INFINIQ}
\affil[]{\{mjjung, jhkim\}@infiniq.co.kr}

\date{}

\renewcommand{\shorttitle}{\textit{arXiv} Template}

\hypersetup{
pdftitle={A template for the arxiv style},
pdfsubject={q-bio.NC, q-bio.QM},
pdfauthor={David S.~Hippocampus, Elias D.~Striatum},
pdfkeywords={First keyword, Second keyword, More},
}

\begin{document}
\maketitle

\begin{abstract}
Few-shot object detection, which focuses on detecting novel objects with few labels, is an emerging challenge in the community.
Recent studies show that adapting a pre-trained model or modified loss function can improve performance.
In this paper, we explore leveraging the power of Contrastive Language-Image Pre-training (CLIP) and hard negative classification loss in low data setting.
Specifically, we propose Re-scoring using Image-language Similarity for Few-shot object detection (RISF) which extends Faster R-CNN by introducing Calibration Module using CLIP (CM-CLIP) and Background Negative Re-scale Loss (BNRL).
The former adapts CLIP, which performs zero-shot classification, to re-score the classification scores of a detector using image-class similarities,
the latter is modified classification loss considering the punishment for fake backgrounds as well as confusing categories on a generalized few-shot object detection dataset.
Extensive experiments on MS-COCO and PASCAL VOC show that the proposed RISF substantially outperforms the state-of-the-art approaches.
The code will be available.
\end{abstract}

\section{Introduction}
In recent years, deep learning has made transformative progress in computer vision areas such as image classification \cite{resnet, vit} and object detection \cite{faster, fast, cascade, detr, retina, yolo}. 
However, a drawback in these fields is that they generally require a large amount of data.
It usually takes much time as well as costs to obtain large amounts of annotated training data.
There are even cases where it is difficult to obtain the training data itself due to security or scarcity of the data.
In contrast, humans excel at generalizing existing knowledge and can quickly understand new concepts based on few examples. 
Therefore, the task of enabling machines to acquire recognition ability with only a small amount of data, similar to humans, is emerging as an important challenge.
For that reason, few-shot learning is receiving a lot of attention.

The goal of few-shot learning is enabling model to predict accurately with only few examples of new classes in training. while early research in few-shot visual learning mainly focused on classification \cite{siam, closer, metasgd,matching} and recognition tasks, it has recently been applied to object detection tasks (FSOD), which is more complex. Initially, FSOD researchers \cite{metayolo, metarcnn} followed the meta-learning paradigm, which resembled solutions used in few-shot classification. However, these methods require complex architecture such as episode settings or new branches. In contrast, transfer-learning paradigms have shown better performance by fine-tuning only the final output layer without the need for complex processes \cite{defrcn,tfa,dcf}. 
In this work, we investigate the re-scoring method using the Contrastive Language-Image Pre-training (CLIP) \cite{clip} top on the transfer-learning paradigm in FSOD.
In addition, we explore performance reduction caused by missing annotations and overcome this issue by modifying the classification loss function

Recently, models pre-trained on publicly available massive datasets have achieved state-of-the-art results in various downstream tasks \cite{beit,eva}. 
In particular, vision and language pretraining approaches have garnered explosive interest \cite{clip, vilbert}. 
However, to the best of our knowledge, research to integrate vision and language pretraining approaches with FSOD has yet to be explored, with most studies utilizing ImageNet-pretrained ResNet as the backbone network \cite{tfa, defrcn, dcf}.
In FSOD, due to the substantial variation performance depending on the sampled images, the common evaluation protocol involves multiple runs using several seeds obtained by k-shot sampling from MS-COCO or Pascal VOC datasets. The results are then averaged to mitigate the impact of this variability.
However, randomly sampled annotation degrades the performance of the detector by causing the missing label issue.

\begin{figure}[!t]
    \centering
    \includegraphics[width=0.7\columnwidth]{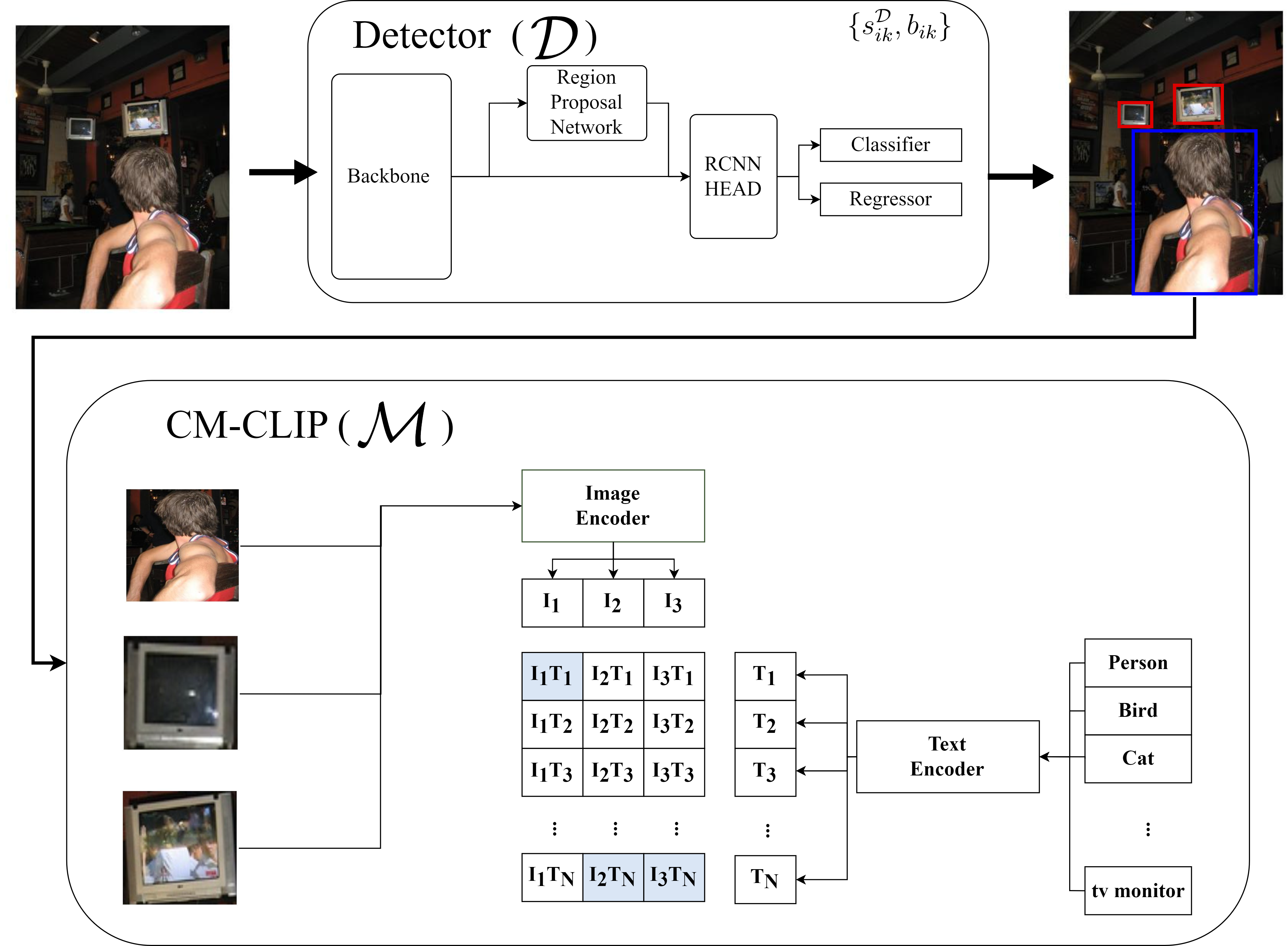}
    \caption{CM-CLIP process scenario. \textbf{Up}: Few-shot object detector predicts bounding boxes and classification scores of object in input image. \textbf{Down}: utilizing pretrained CLIP model, CM-CLIP re-inference cropped images obtained from predicted bounding boxes. The resulting re-inferred output is subsequently integrated with the detector's inference outcome to yield the final score.}
    \label{fig:cmclip0}
\end{figure}

Motivated by the above observations, we propose a Re-scoring using Image-Language Similarity for FSOD (RISF) containing Calibration Module using CLIP (CM-CLIP) and Background Negative Re-scale Loss (BNRL).
The former utilizes CLIP pre-trained on an open dataset of 400M (image, text) pairs to calibrate the scores predicted by the detector. The proposed module aims to enhance detection performance by leveraging the pre-trained CLIP model.
The latter is a modified focal loss \cite{retina} as the loss function for RCNN classification.
Specifically, CM-CLIP uses a pre-trained CLIP model that enables zero-shot classification through the similarity between image and text, as shown in Figure \ref{fig:cmclip0}. 
Thus, it can supplement the detector's deficient classification power by calculating the similarity between the predicted object area of the detector and the class name.
Furthermore, BNRL is designed based on focal loss that considers the punishment of the model's prediction of background areas and hard negative samples.
While CM-CLIP may be confused with the concept of Open Vocabulary Object Detection \cite{vild,regionclip}, it infers instances only from a limited set of known classes and only adjusts the score of the region predicted by the few-shot detector. 

The main contributions of our work are summarized as follows:

\begin{itemize}
    \item To the best of our knowledge, to re-calculate the classification score of the detector, we are the earlier study in FSOD to utilize the similarity score between image and class name using CLIP.
    \item We explore a missing annotation problem that compels a model to learn object area as background and causes the model to be confused during training. To adjust such a problem, we propose modified classification loss.
    \item Extensive experiments on generalized few-shot object detection of MS-COCO and PASCAL VOC demonstrate that our approaches substantially improve the state-of-the-art.
\end{itemize}

\section{Related Works}

\subsection{Few-shot learning}
Few-shot learning aims to learn new concepts only providing a few labeled examples. In the field of few-shot classification, methods based on meta-learning \cite{metasurvey, maml, metasgd,latent} and metric-learning approaches \cite{siam, matching, dynamic} are primarily proposed. Meta-learning-based approaches focus on learning a meta-model on episodes of tasks to adapt novel tasks with few examples, known as ``learning-to-lean'', 
while metric-based approaches attach importance to learning a good feature representation embedding space that can be used in downstream tasks. 
Outside of meta-learning and metric learning, fine-tuning-based approaches \cite{closer, rethink} achieve competitive performance by fine-tuning linear classifiers on top of pre-trained models.
Compared to classification context, few-shot object detection involves localization and is even more challenging due to the problem of missing annotations caused by a small amount of annotations.

\subsection{Few-Shot Object Detection}
Few-shot object detection can be classified as meta-learning paradigm \cite{metayolo, metadetr, metarcnn, dana, metadet, fsview} and fine-tuning paradigm \cite{tfa,defrcn,dcf, fssp}. The meta-learning based approach mainly uses feature aggregation through multiple episodes for a support set of query images. FSRW \cite{metayolo} based on YOLO v2 \cite{yolov2} reweights features through channel-wise multiplication. DANA \cite{dana} uses two attention blocks, namely background attention block and cross-image spatial attention block, to undermine background features and aggregate query support images through QKV attention. Meta-DETR \cite{metadetr} uses an attention mechanism based on intraclass correlation, leveraging DETR \cite{detr}. The fine-tuning based approach, first proposed by TFA \cite{tfa}, focuses on fine-tuning Faster RCNN \cite{faster} trained on base classes for novel classes. During fine-tuning, TFA randomly initializes output layers and freezes the rest of the components. However, in DeFRCN \cite{defrcn}, RPN and backbone are unfrozen, and backbone is trained with controlled gradient flow, resulting in improved performance. DCF \cite{dcf} focuses on mislabeling in few-shot datasets by creating a background head, allowing the detector to learn mislabeling.

\subsection{Joint Vision and Language Modeling}
In recent years, there have been significant advances in the field of vision-language cross-modal tasks such as open vocabulary object detection \cite{openvocab, vild, regionclip} and language-based visual retrieval \cite{videoclip,imageret}. 
Among them, CLIP \cite{clip} learns a multi-modal embedding space using a dataset of 400M image-text pairs, by incorporating the cosine similarity to be maximized for correct paris and minimized for incorrect pairs.
CLIP has demonstrated powerful zero-shot performance on variety of datasets.
CLIP has been applied in various fields. In the text-to-image generation domain \cite{styleclip, dalle}, it is used to find an image embedding that is located in the same position as the text in the pre-trained embedding space to generate an image related to the text. In addition, it can be utilized in open vocabulary object detection \cite{regionclip} to find a vocabulary with an embedding identical to the detected object.
In this paper, we leverage the power of CLIP to retrieve similarity scores between image embedding and text embedding in FSOD.

\section{Methods}

\subsection{Problem definition}
In this paper, we consider the standard problem setup as previous few-shot object detection (FSOD) works \cite{tfa, defrcn, dcf}.
Specifically, the FSOD dataset is constructed by partitioning a general object detection dataset ${D} = \{ {X}_i, {Y}_i\}^{N}_{i=1}$, where $X_i$ refer to the i-th image and $Y_i$ is its corresponding bounding box and class ${Y}_i=\{b_k,c_k\}_k^M$, into a non-overlapping novel and base classes, denoted as $C_{novel} \cap C_{base} = \varnothing$.
The novel set comprises novel classes with only K (e.g. 1,5, and 10) instances per category, while the base set contains abundant instances per category of base classes.
As mentioned above, we build an RISF upon the transfer learning paradigm, which involves two stages, i.e., base training on the base set and fine-tuning it on the novel set. 
In addition to what we mainly propose, the RISF also uses the Gradient Decoupling Layer proposed by Quao et al \cite{defrcn} for a strong baseline.
We evaluated our model in two common protocols, including few-shot object detection (FSOD) and generalized few-shot object detection (gFSOD). 
FSOD setting evaluates the model by training only on novel classes during fine-tuning stage.
However, in the gFSOD setting, the novel training set is composed of an equal balance of base and novel classes.

\subsection{CM-CLIP}
\begin{algorithm}[!b]
\SetAlgoLined

    \PyComment{$\mathcal{M}_{image}$ = CLIP.image\_encoder } \\
    \PyComment{$\mathcal{M}_{text}$ = CLIP.text\_encoder} \\
    \PyComment{$\mathbf{W}_{image}$ = CLIP.image\_proj\_layer} \\
    \PyComment{$\mathbf{W}_{text}$ = CLIP.text\_proj\_layer} \\
    \PyComment{$\mathcal{F}$ = few-shot detector} \\
    \PyComment{$\mathcal{R}$ = ROI Pooler} \\
    
    \PyComment{I[n,h,w,3] = RGB images on test datasets} \\
    \PyComment{T[K,l] = k-th class names} \\
    
    \PyCode{}\\
    \PyComment{class name to text embedding T[K,l] $\rightarrow$ $T_e$[K,$d_e$]} \\
    \PyCode{$ \mathbf{T\_f} = \mathcal{M}_{text}(\mathbf{T})$} \\
    \PyCode{$ \mathbf{T\_e} = \text{normalize} (\mathbf{W}_{text} \cdot \mathbf{T\_f}$}) \\
    
    \PyCode{}\\
    \PyCode{for $I_n$ in I:}\\
    \Indp
        \PyComment{$ \mathbf{s}^\mathcal{F}, \mathbf{b} = \{s^\mathcal{F}_i, b_i \  | \text{ i is predicted object index on a single image, and } 1 < i < M \  \}$ }\\
        \PyCode{$(\mathbf{s}^\mathcal{F}, \mathbf{b})_n$ = $\mathcal{F}(I_n)$}\\
        \PyCode{}\\
        
        \PyComment{crop the image according to predicted bounding box, $\mathbf{c}_n$[$M$,h',w',3]}\\
        \PyCode{$\mathbf{c}_n$ = $\mathcal{R}(I_n, \mathbf{b}_n)$}\\
        \PyCode{}\\
        
        \PyComment{cropped images to image embeddings, $\mathbf{c}_n$[$M$,h',w',3] $\rightarrow$ $\mathbf{I\_e}$[$M$,$d_e$]}\\
        \PyCode{$\mathbf{I\_f}$ = $\mathcal{M}_{image}(\mathbf{c}_n)$}\\
        \PyCode{$ \mathbf{I\_e} = \text{normalize} (\mathbf{W}_{image} \cdot \mathbf{I\_f}$}) \\
        \PyCode{}\\
        
        \PyComment{Calculate similarity score $\mathbf{s}^\mathcal{M}$[$M$,K] }\\
        \PyCode{ $\mathbf{s}^\mathcal{M} =  \text{normalize(}{\exp(\mathbf{I\_e} \cdot \mathbf{T\_e^\mathsf{T}} \slash \tau )} \text{, dim = 1)}$ }\\
        \PyCode{}\\

        \PyComment{re-scoring }\\
        
        \PyCode{$\mathbf{S} = \alpha \mathbf{s}^\mathcal{F} + (1-\alpha) \mathbf{s}^\mathcal{M}$}
        
    \Indm
    
\caption{Pseudocode for the core of CM-CLIP}
\label{algo:your-algo}
\end{algorithm}

In this section, we introduce a classification score re-scoring method called CM-CLIP, which stands for Calibration Module using CLIP. 
This method improves the classification performance of few-shot detectors trained on only limited training data by utilizing CLIP.
Recent advances in joint vision and language pre-trained models, such as CLIP \cite{clip}, have shown powerful zero-shot transfer performance on a variety of downstream tasks.
CLIP learns multi-modal embedding space using contrastive learning, which maximizes the cosine similarity between correct text-image pairs and minimizes incorrect pairs. It is trained using 400M  image and text pairs and can perform zero-shot classification on unseen images and class names by determining which text-image pair is more probable. 
CM-CLIP adapts this zero-shot image classification capability by calculating the similarity between the cropped images and the corresponding class name without additional training.

Figure \ref{fig:cmclip0} shows the re-scoring process using CM-CLIP for the classification score of a few-shot detector at test time.
As shown in Figure \ref{fig:cmclip0} above, the detector predicts the bounding boxes and classification scores of the objects in the image. 
Then we can crop object images using predicted bounding boxes. As shown in Figure \ref{fig:cmclip0} below, CM-CLIP converts $i$-th cropped images and $k$-th class name into embeddings as $\mathbf{I\_e} = \{\text{I\_e}_1,...,\text{I\_e}_M\}, \mathbf{T\_e} = \{ \text{T\_e}_1,..., \text{T\_e}_K\}$ using the pre-trained image-text encoders of CLIP.
If the i-th object region resembles a human shape, then this region and the text ``person" are likely to be located in a similar position in the multi-modal embedding space.
Finally, CM-CLIP calculates similarity scores between the class name (e.g. person, bird, cat,..., and tv monitor) and object area in embedding space.
The similarity between images and classes is estimated by  
\begin{equation}
    {s}^{\mathcal{M}} =  \frac{\exp( \mathbf{I\_e} \cdot \mathbf{T\_e} \slash \tau  )}{\sum_k \exp(\mathbf{I\_e} \cdot \mathbf{T\_e} \slash \tau )  }
\end{equation}
where $i$ and $k$ denote the indices of the image and class, respectively, while $\tau$ refers to temperature scaling. 
This similarity score is subsequently used to re-score the classification score of the few-shot detector.  
The CM-CLIP then computes the final class score by weighted averaging scores of the detector and CLIP:
\begin{equation}
    \mathbf{S} = \mathbf{c} \mathbf{s}^\mathcal{F} + (1-\mathbf{c}) \mathbf{s}^\mathcal{M}
    \label{eq:cm-clip}
\end{equation}
where $\mathbf{c}$ is a  hyperparameter for regulating the importance of each score.

The CM-CLIP extends the PCB (Prototypical Calibration Block) \cite{defrcn} which is pre-trained with classification on Imagenet1k \cite{IMAGENET} and generates prototypes of classes by averaging few ROI features. In contrast, CM-CLIP was pre-trained with image-text contrastive learning on LAION-400M \cite{LAION} and utilized text embeddings as prototypes for classes. 
These enhancements have contributed significantly to the notable performance difference between RISF and DeFRCN.

\subsection{BNRL}
\begin{figure}[ht]
    \centering
    \begin{subfigure}[]{0.45\textwidth}
    \includegraphics[width =\textwidth,height=0.7\textwidth]{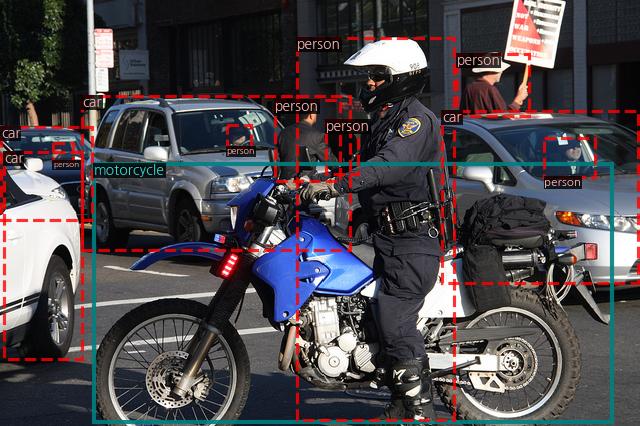}
    \caption{an image on seed with randomly sampled annotations}
    \end{subfigure}
    \begin{subfigure}[]{0.48\textwidth}
\includegraphics[width=\textwidth]{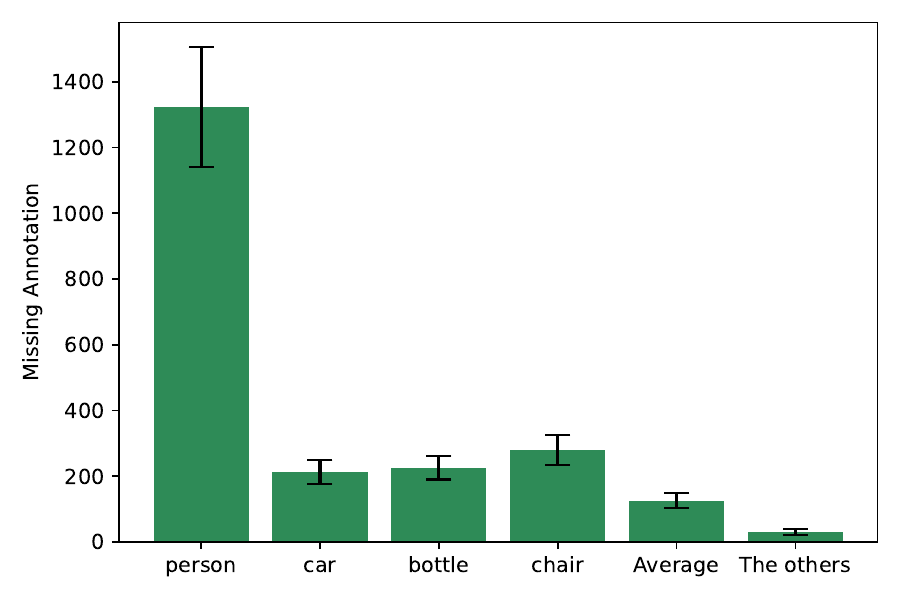}
    \caption{The number of missing annotations on seed with randomly sampled annotations}
    \end{subfigure}
    \caption{
The missing annotation on FSOD protocol dataset. 
(a) shows an empirical example of missing annotations on MS-COCO in FSOD protocol indicated by the red-dotted boxes.
In (b), we present the number of missing annotations on MS-COCO 10-shot training dataset in FSOD protocol.}
    \label{fig:negative sample1}
\end{figure}
In this section, first, we explain the definition of missing annotation, why it is inevitable in FSOD protocols, and how much it degrades the performance of the detector.
Next, we introduce Background Negative Re-scaling Loss (BNRL) which alleviates the missing annotation and hard negative.

In general FSOD evaluation protocol \cite{tfa}, models are evaluated as an average of performance with multiple seeds sampled from general object detection datasets (e.g. MS-COCO).
In most of the seeds, a very large amount of objects are not labeled, as observed in Figure \ref{fig:negative sample1}(a), because annotations are randomly sampled.
Therefore the person and many others dotted red boxes in Figure\ref{fig:negative sample1} (a) will be regarded as background during few-shot fine-tuning.
We define the seeds containing these randomly sampled objects as random seeds, and the objects whose annotations have disappeared through this process as missing annotations.

Figure \ref{fig:negative sample1} (b) describes the average and 95\% confidence of missing annotations on MS-COCO 10-shot.
The classes ``person'', ``cat'', ``bottle'', and ``chair'' represent the four classes with the highest number of missing annotations. The ``average'' indicates the average value across all 20 novel classes, while ``the others'' represents the average value of the remaining 16 classes.

\begin{figure}[ht]
    \centering
    \begin{subfigure}[]{0.45\textwidth}
    \includegraphics[width =\textwidth,height=0.7\textwidth]{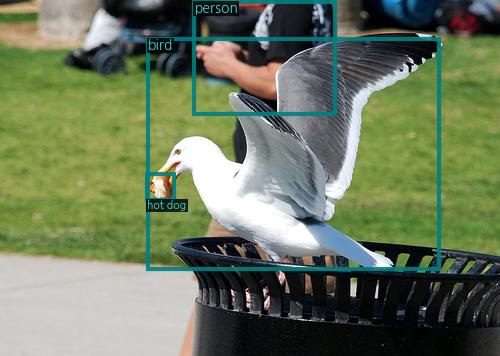}
    \caption{an image on seed with carefully selected annotations}
    \end{subfigure}
    \begin{subfigure}[]{0.48\textwidth}
\includegraphics[width=\textwidth]{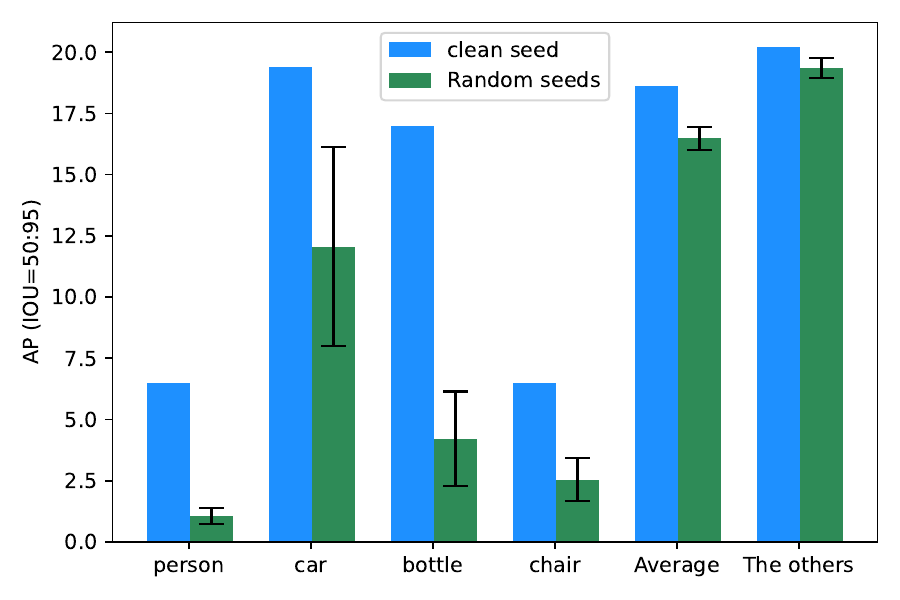}
    \caption{Average precisions of DeFRCN trained on clean seed and random seeds.}
    \end{subfigure}
    \caption{
The clean annotations on FSOD protocol dataset. 
(a) shows an empirical example of a specific seed with carefully selected annotations on MS-COCO in FSOD protocol.
In (b), the model trained on random seeds with such missing annotations exhibits lower AP compared to the model trained on clean seed with very few missing annotations, which are obtained by carefully selected images. For the error bar, we report the average and 95\% confidence.}
\label{fig:negative sample2}
\end{figure}
    
However, very specific seeds are carefully selected, and thus these seeds rarely have missing annotation as observed in Figure \ref{fig:negative sample2}(a).
In this paper, we define these seeds as clean seeds.
Figure \ref{fig:negative sample2} (b) illustrates the performance difference in terms of AP (IOU=50:95) between the clean seed (blue) and the remaining random seeds (green) as measured by the DeFRCN model.
In cases where there were many missing annotations, there was a significant drop in performance. However, in the remaining classes with almost no missing annotations, there was little to no drop in performance.
Therefore, missing annotations degrade performance for even a two-stage detector.
We hypothesize that the use of cross entropy in FSOD learns the noise as it is, so we should introduce a loss term to prevent this.

The issue of missing annotations has been previously tackled by Zhang et al \cite{BRL}. 
Their approach mitigates this problem through a transition from the conventional loss function to the loss associated with the mirrored positive branch when the confidence score falls below the threshold.
However, it didn't consider that there are many types of hard negatives that can occur in FSOD protocols (e.g. cat and dog).
Xu et al. \cite{fssp} also proposed a loss that takes into account hard negative examples due to similar categories, but this loss is vulnerable the noisy annotations.
Therefore, we introduce BNRL, which stands for Background Negative Re-scaling Loss. This loss focuses more on hard negative samples due to similar shapes, but less on the background:

\begin{equation}
    \mathcal{L}_c = 
            - 
            \beta \hat{p}(c)(1-p(c))^\gamma \log(p(c)) 
            -
            (1-\beta)(1-\hat{p}(c)) (p(c))^\epsilon \log(1-p(c)) 
    \label{eq:BNRL1}
\end{equation}

\begin{equation}
    \text{BNRL} = \sum_{\{bg \notin C\}}{\mathcal{L}_c} + \omega_{bg}\mathcal{L}_{bg}
    \label{eq:BNRL2}
\end{equation}

where $\gamma$ and $\epsilon$ are the scaling-factors for loss, 
$\beta$ is a balancing parameter between positive and negative terms and $\omega_{bg}$ is a weight for background.
In addition, $\hat{p}(c)$ represents the distribution of the ground truth class. For instance, in the case of a dog object, $\hat{p}(dog)$ =1 and $\hat{p}(cow)$ = 0.
In this change, our detector can more concentrate on hard negative foreground objects and less concentrate on the background.

The $\gamma$ and $\epsilon$ in BNRL represent the factors that incorporate the easy negative factor into the focal loss. On the other hand, the other two hyperparameters ($\beta$ and $\omega_{bg}$) are introduced to overcome the challenges of the FSOD setting. The $\beta$ is related to the mirror term effective for hard negative, while $\omega_{bg}$ is a weight term for the background effective for missing annotations.

\begin{figure}[h]
    \centering
    \includegraphics[width=0.65\linewidth]{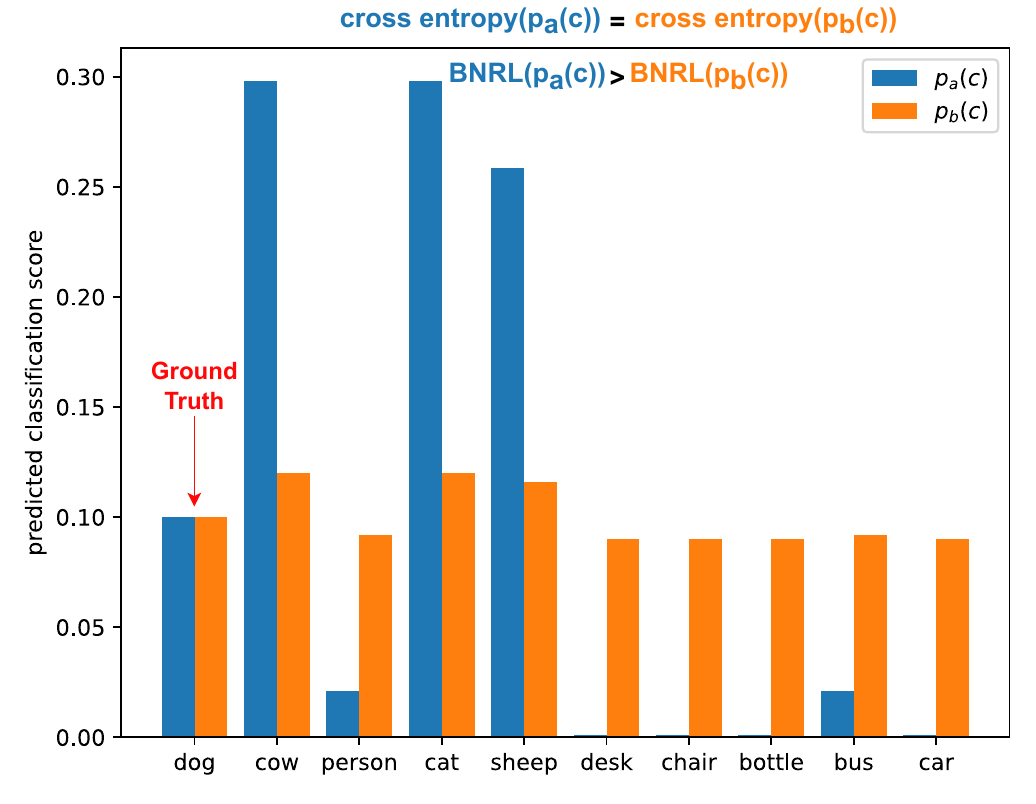}
\caption{Two classification score distributions. When the ground truth is "dog," the cross-entropy or focal loss of the two classification score (orange and blue) are the same. However, BNRL shows that the orange distribution is larger than the blue distribution. $\alpha$ is the scaling factor of the noise.
}
\label{fig:BNRLvsCROSS}
\end{figure}

Cross-entropy is calculated based on the classification score for the ground truth. 
Therefore, it does not reflect whether the detector predicts all classes with the same score or strongly misclassifies certain classes.
However, the mirror term of BNRL is designed to take a larger loss when the model significantly misclassifies (resulting in high classification scores but incorrect classifications) compared to cases with generally low classification scores as shown in Figure \ref{fig:BNRLvsCROSS}.

To verify how BNRL punishes hard-negative examples, we consider $p_\alpha(c)$ as a hypothetical probability function whose detector predicts incorrect class c.

\begin{equation} 
p_\alpha(c) = \frac{ \frac{1}{N} + \alpha x_{c}}{Z_a}
\label{eq:p}
\end{equation}

where $Z$ is the normalization factor, $N$ is the number of incorrect classes, $x_{c}$ is the random noise term with a sum of zero for class c (i.e. $\sum_c{x_{c}} = 0$), and $\alpha$ is the scaling constant for noise.
This distribution indicates that when $\alpha$ is large, the model is highly confident about certain incorrect classes (i.e. hard negative) as shown in Figure \ref{fig:BNRLvsCROSS} blue. On the other hand, when $\alpha$ is small, the model poorly predicts all classes uniformly as shown in Figure \ref{fig:BNRLvsCROSS} orange.

If the following equation holds true, the mirror term is a monotonic increase for $\alpha$.
In other words, the BNRL more concentrates on hard negative examples than easy negative examples.

\begin{equation} - \sum_c \log (1-p_{\alpha_1}(c)) \leq - \sum_c \log (1-p_{\alpha_2}(c)) \quad \text{when} \quad \alpha_1 < \alpha_2 \end{equation}

{\it proof.} 
First, we simplify the $Z$ of Equation \ref{eq:p}.

\begin{equation}
\begin{split}
    \text{then}, \quad Z &= \sum_c \left[ \frac{1}{N} + \alpha x_{c} \right] \\
    &= \sum_c \left[ \frac{1}{N} \right] + 0  \\
    &= {\text constant}
\end{split}
\end{equation}

Next, we define $g(x_i) = \log \frac{1- (\alpha_2 x_i + \frac{1}{N})}{1-(\alpha_1 x_i + \frac{1}{N})}$.
When $\alpha_1 < \alpha_2$, $g(x_i)$ is concave, and according to Jensen's inequality, 

\begin{equation}
\begin{split}
    &\sum_c g(x_c) \leq g\left(\sum_c x_c\right).\\
    & \Rightarrow  \sum_c g(x_c) \leq g(0)\\
    & \Rightarrow \sum_c g(x_c) \leq 0 \\
    & \Rightarrow  \sum_c \left[ \log \frac{1 - (\alpha_2 x_{c} + \frac{1}{N})}{1 - (\alpha_1 x_{c} + \frac{1}{N})} \right] \leq 0 \\
    & \Rightarrow  \sum_c \left[ \log\left(1 - \left({\alpha_2 x_{c}+\frac{1}{N}}\right)\right) \right]   \leq \sum_c \left[ \log\left(1 - \left({\alpha_1 x_{c} + \frac{1}{N} }\right)\right)\right]  \\
    & \Rightarrow - \sum_c \log (1 -  p_{\alpha_1}(c)) \leq - \sum_c \log (1 - p_{\alpha_2}(c))
\end{split}    
\label{eq:proove}
\end{equation}

Thus, Equation \ref{eq:proove} shows $- \sum_c \log \left( 1- p_\alpha(c) \right)$ is a monatonic increase for $\alpha$.

\begin{figure}[h]
    \centering
    \includegraphics[width=0.65\linewidth]{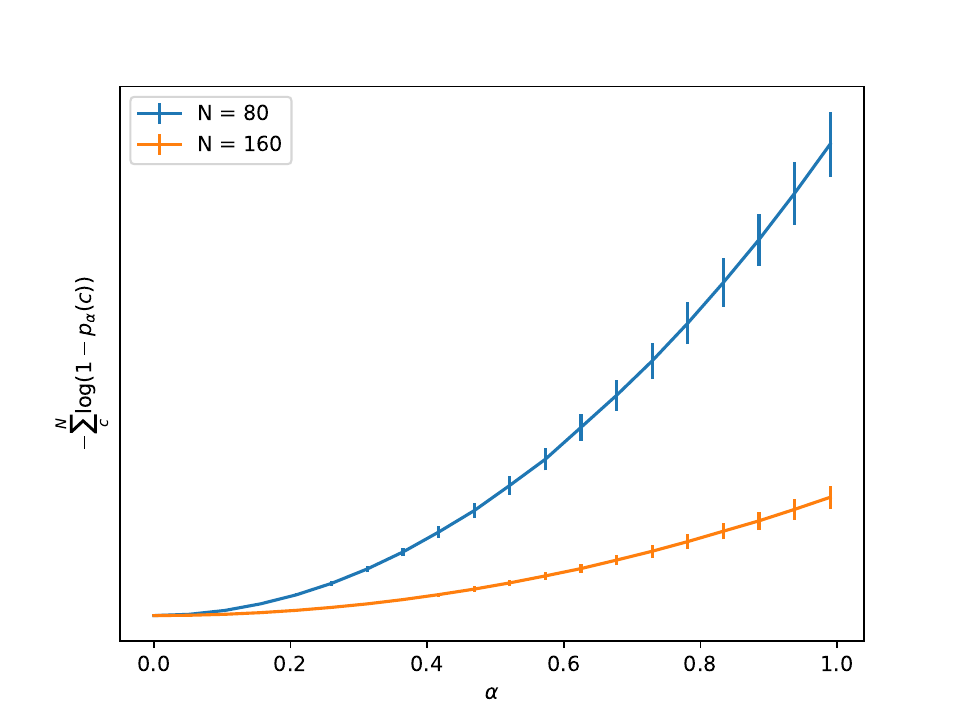}
\caption{When $p_\alpha \propto \frac{1}{N} + \alpha x_{c}$, the graph of the mirror term is $- \sum \log(1-p(c))$. The mirror term is a monotonically increasing function depending on $\alpha$. 
}
\label{fig:numerical}
\end{figure}

We have confirmed these results numerically in Figure \ref{fig:numerical}. This loss design assumes scenarios where the model retains strong memorization of base classes, making it challenging to accommodate novel classes. 
Such cases are classified as hard negative and thus can be alleviated by using the mirror term of BNRL.
However, the mirror term becomes vulnerable when the detector predicts accurate foreground objects with high classification scores, but there are missing annotations for those instances. To address this issue, we introduced a background weight $\omega_{bg}$ to reduce the model's focus on learning backgrounds. 
As a result, BNRL has an advantage in the setting where hard negative and missing annotations exist.

\medskip

\section{Experiments}
In this section, we maintain adherence to the established evaluation protocol \cite{tfa} and undertake a comparative analysis between our model and contemporary state-of-the-art approaches on the benchmark that has been previously employed in scholarly investigations.

\subsection{Experimental Setup}
To ensure a fair and meaningful comparison with prior models, we evaluate our model using the widely adopted MS-COCO \cite{coco} and Pascal VOC \cite{voc} k-shot settings.

{\bf MS-COCO} consists of a total of 80 categories, with 60 base categories and 20 novel categories. We trained the novel categories in a k-shot setting and evaluated the AP (IOU=50:95) of novel categories using the MS-COCO validation dataset. 
As mentioned above, there is a significant amount of missing annotations for the MS-COCO k-shot setting.

{\bf Pascal VOC} setting in few-shot object detection consists of a total of 20 object categories from Pascal VOC 07+12, divided into 15 base object categories and 5 novel object categories. For the few-shot training, the novel categories have K=1,2,3,5, and 10 instances per class. Consistent with prior research, we divide the categories into three sets, namely novel sets 1, 2, and 3. For evaluation, we report AP (IOU=50) for novel classes based on the test set of Pascal VOC 2007.
In the k-shot setting of Pascal VOC, the instances are not as dense, so there are fewer missing annotations compared to the MS-COCO k-shot setting.

{\bf Implementation details.} Our approach is conducted with Detectron2 \cite{detectron2} for detection framework on 8 NVIDIA A100 on CUDA 11.1. 
We employ Faster R-CNN with Gradient Decoupling Layer (GDL) for our main detector, which is the same architecture as DeFRCN \cite{defrcn}. 
We optimize our model using SGD with a minibatch size of 16, a momentum of 0.9, a weight decay of $5e^-5$, and a learning rate of 0.2 during the base training and 0.01 during the fine-tuning stage.
We use the BRNL only during few-shot fine-tuning with $\beta=0.2, \gamma=4.0, \epsilon=1.0, \omega_{bg}=0.2$, and the CM-CLIP with $\mathbf{c}=0.7$ and $\tau=0.01$. 
The pretrained ViT-L/14@336px is used as the vision encoder of CM-CLIP.

Following the two-stage paradigm \cite{tfa}, we train a base model using a large amount of labeled data from base classes in the first stage, and in the second stage, 
we randomly initialize output layers of the base model and then fine-tune this model using a small amount of data from novel classes.
The base model of RISF is identical to DeFRCN.
In other words, we do not utilize our approach during the first stage.
Therefore, we only use BNRL during the fine-tuning stage, and CM-CLIP is used during testing. 
To ensure a fair comparison, the structure and all hyperparameters, such as learning rate, batch size, and $\lambda$ in GDL, of the detector remain the same as in DeFRCN \cite{defrcn}.

\subsection{Comparison with State-of-the-Art}

\begin{table}[h]
\caption{Few-shot object detection performances for different k-shot, with and without generalized few-shot object detection(gFSOD) settings on MS-COCO.
The results are averaged over all 10 seeds and
the boldface indicates the best.
The underline $\underline{\mathbf{x}}$ indicate that the results were reported in DeFRCN \cite{defrcn}. $-$ indicates no results reported in papers.
}

  \centering
  \begin{adjustbox}{width=1.0\textwidth}
  {
  \begin{tabular}{l|c|cc|cc|cc|cc|cc|cc}
    \toprule
    \multirow{2}{*}{Method \slash shot}& \multirow{2}{*}{w \slash G}& \multicolumn{2}{c|}{1}& \multicolumn{2}{c|}{2}& \multicolumn{2}{c|}{3}& \multicolumn{2}{c|}{5}& \multicolumn{2}{c|}{10}& \multicolumn{2}{c}{30}\\
     &&  AP &AP50&AP&AP50&AP&AP50&AP&AP50&AP&AP50&AP&AP50\\
    \hline
    FRCN-ft \cite{metarcnn} &\XSolidBrush& $\underline{1.0}$ & - & $\underline{1.8}$ & - & $\underline{2.8}$ & - & $\underline{4.0}$ & - & 6.5 & - & 11.1& - \\
    TFA \cite{tfa} &\XSolidBrush& $\underline{4.4}$ & - & $\underline{5.4}$ & - & $\underline{6.0}$ & - & $\underline{7.7}$ & - & 10.0 & - & 13.7& - \\
    MPSR \cite{mpsr} &\XSolidBrush& $\underline{5.1}$ & - & $\underline{6.7}$ & - & $\underline{7.4}$ & - & $\underline{8.7}$ & - & 9.8 & - & 14.1& - \\
    FSDetView \cite{fsview} &\XSolidBrush& 4.5 & - & 6.6 & - & 7.2 & - & 10.7 & - & 12.5 & - & 14.7& - \\
    QA-FewDet \cite{QA} &\XSolidBrush& 4.9&- &7.6&-&8.4&-&9.7&-&11.6&-&16.5&-\\
    FADI \cite{FADI}&\XSolidBrush&5.7 & - & 7.0&-&8.6&-&10.1&-&12.2&-&16.1&-\\
    DeFRCN \cite{defrcn} &\XSolidBrush& 7.7&15.1&11.4&21.4&13.3&24.5&15.5&28.3&18.5&33.4&22.5&39.5 \\
    DCF \cite{dcf} & \XSolidBrush& 8.1&16.3&12.1&23.4&14.4&27.1&16.6&31.1&19.5&35.8&22.7&41.0\\
    MFDC \cite{MFD}&\XSolidBrush& 10.8&-&13.9&-&15.0&-&16.4&-&19.4&-&22.7&-\\
    \rowcolor{Gray}
     Ours&\XSolidBrush& {\bf 11.7}&{\bf 24.6}&{\bf 15.9}&{\bf 31.5}&{\bf 18.2}&{\bf 34.8}&{\bf 20.3}&{\bf 38.0}&{\bf 21.9}&{\bf 39.9}&{\bf 24.4}&{\bf 43.2}\\
    \midrule
    FRCN-ft \cite{metarcnn}&\checkmark&1.7&3.3&3.1&6.1&3.7&7.1&4.6&8.7&5.5&10.0&7.4&13.1\\
    TFA \cite{tfa}&\checkmark&1.9&3.8&3.9&7.8&5.1&9.9&7.0&13.3&9.1&17.1&12.1&22.0\\
    DeFRCN \cite{defrcn}&\checkmark&4.8&9.5&8.5&16.3&10.7&20.0&13.6&24.7&16.8&29.6&21.2&36.7\\
    DCF \cite{dcf} &\checkmark&6.2&12.7&10.4&20.9&12.9&25.1&15.7&30.3&18.3&34.5&21.9&{\bf 40.2}\\
    \rowcolor{Gray}
    Ours&\checkmark&{\bf 9.2}&{\bf 18.6}&{\bf 13.4}&{\bf 26.1}&{\bf 15.6}&{\bf 29.6}&{\bf 17.8}&{\bf 33.2}&{\bf 20.5}&{\bf 37.4}&{\bf 22.5}&{\bf 40.2}\\
    \bottomrule

  \end{tabular}
  }
\end{adjustbox}
\label{tab:MS-COCO}
\end{table}

\begin{table*}[!hbt]
\caption{Few-shot object detection performances of novel classes (nAP50) on Pascal VOC.
We evaluate using the gFSOD setting with three different splits. The superscript $^{*}$ indicates that the base class is not learned on the fine-tuning stage (Equal to $\text{wo} \slash \text{G}$).}
  \centering
\begin{adjustbox}{width=1\textwidth}
  {
  \begin{tabular}{l|ccccc|ccccc|ccccc}
    \toprule
    \multirow{2}{*}{Approach} & \multicolumn{5}{c| }{Novel Set 1}& \multicolumn{5}{ c |}{Novel Set 2}& \multicolumn{5}{ c }{Novel Set 3}   \\
     &   K = 1 & 2 & 3 & 5 & 10 & K = 1 & 2 & 3 & 5 & 10 & K = 1 & 2 & 3 & 5 & 10    \\
    \hline
     FRCN+ft \cite{metarcnn}  & 15.2 &20.3 & 29.0 & 25.5 & 28.9 &13.4& 20.6 &28.6& 32.4 & 38.8 &19.6&20.8&28.7&42.2&42.1   \\
     FSRW \cite{metayolo}& 14.8 &15.5 &26.7 & 33.9 & 47.2 & 15.7 & 15.3 & 22.7 & 30.1 & 39.2 & 19.2 & 21.7 & 25.7 & 40.6 & 41.3 \\
    Meta Det \cite{metadet}  & 18.9&20.6&30.2&36.8&49.6&21.8&23.1&27.8&31.7&43.0&20.6&23.9&29.4&43.9&44.1\\
    TFA w$\backslash$fc \cite{tfa}  & 36.8 &29.1 & 43.6 & 55.7 & 57.0 &18.2& 29.0 &33.4& 35.5 & 39.0 & 27.7&33.6&42.5&48.7&50.2  \\
    TFA w$\backslash$cos \cite{tfa} & 39.8 &36.1 & 44.7 & 55.7 & 56.0 &23.5& 26.9 &34.1& 35.1 & 39.1 & 30.8&34.8&42.8&49.5&49.8  \\
    MPSR \cite{mpsr} & 42.8 &43.6&48.4&55.3&61.2&29.8&28.1&41.6&43.2&47.0&35.9&40.0&43.7&48.9&51.3\\
     MM-FSOD \cite{MM} & 42.5 &41.2 & 41.6 & 48.0 & 53.4 &30.5& 34.0 &39.3& 36.8 & 37.6 &39.9&37.0&38.2&42.5&45.6  \\
     QA-FewDet \cite{QA} & 42.4 &51.9 & 55.7 & 62.6 & 63.4 &25.9& 37.8 &46.6& 48.9 & 51.1 &35.2&42.9&47.8&54.8&53.5  \\
    KFSOD \cite{KFSOD} & 44.6 &- & 54.4 & 60.9 & 65.8 &37.8&-& 43.1 &48.1& 50.4 & 34.8 &-&44.1&52.7&53.9  \\
     TENET \cite{TENET} & 46.7 &- &55.4 & 62.3 & 66.9 & 40.3 &-& 44.7 &49.3& 52.1 & 35.5 &-&46.0&54.4&54.6  \\
     FCT \cite{FCT} & 49.9 & 57.1 & 57.9 & 63.2 & 67.1 &27.6& 34.5 &43.7& 49.2 &51.2& 39.5 &54.7&52.3&57.0&58.7  \\
     FADI\cite{FADI} &50.3&54.8&54.2&59.3&63.2&30.6&35.0&40.3&42.8&48.0&45.7&49.7&49.1&55.0&59.6\\
     ICPE \cite{ICPE} & 54.3 & 59.5 &62.4 & 65.7 & 66.2 & 33.5 &40.1& 48.7 &51.7& 52.5 & 50.9 &53.1&55.3&60.6&60.1  \\
     LVC \cite{LVC} & 54.5 &53.2 & 58.8 & 63.2 & 65.7 &32.8& 29.2 &50.7& 49.8 & 50.6 &48.4&52.7&55.0&59.6&59.6  \\
    DeFRCN \cite{defrcn} &57.0 &58.6&64.3&67.8&67.0&35.8&42.7&51.0&54.4&52.9&52.5&56.6&55.8&60.7&62.5 \\
     CNPB-DeFRCN \cite{CNPB} & 57.2 &- & 63.0 & 66.2 & 66.6 &39.7& - &51.8& 54.7 & 53.1 &51.0&-&56.9&57&60.7  \\
     MFDC \cite{MFD} & 63.4 &66.3 & 67.7 & 69.4 & 68.1 &42.1& 46.5 &53.4& 55.3 & 53.8 &56.1&58.3&59.0&62.2&63.7  \\
    \rowcolor{Gray}
    Ours   &{\bf 67.2}&{\bf 70.5}&{\bf 71.5}&{\bf 74.2}&{\bf 73.9}&{\bf 47.6}&{\bf 52.3}&{\bf 57.3}&{\bf 58.3}&{\bf 60.4}&{\bf 59.4}&{\bf 59.0}&{\bf 59.1}&{\bf 62.4}&{\bf 63.9}\\
    \hline 
    $\text{DCF}^{*}$ \cite{dcf}   &56.6&59.6&62.9&65.6&62.5&29.7&38.7&46.2&48.9&48.1&47.9&51.9&53.3&56.1&59.4 \\
    \rowcolor{Gray}
    $\text{Ours}^{*}$   &{\bf 66.7}&{\bf 70.3}&{\bf 71.5}&{\bf 72.3}&{\bf 72.3}&{\bf 47.5}&{\bf 53.9}&{\bf 59.8}&{\bf 62.1}&{\bf 58.6}&{\bf 61.1}&{\bf 61.9}&{\bf 64.6}&{\bf 65.2}&{\bf 65.9}\\
    \midrule 
\multicolumn{16}{l}{\textit{Results averaged over multiple random runs:}}\\
    FRCN+ft \cite{metarcnn} & 9.9& 15.6  &21.6  &28.0  &52.0  &9.4& 13.8& 17.4&21.9& 39.7 &8.1&13.9&19.0&23.9&44.6\\
    TFA  \cite{tfa} & 25.3 & 36.4 & 42.1 & 47.9 & 52.8 & 18.3 & 27.5 & 30.9 & 34.1 & 39.5 & 17.9 & 27.2 & 34.3 & 40.8 & 49.6 \\
    DeFRCN \cite{defrcn}  &40.2 & 53.6 & 58.2 & 63.6 & 66.5 &29.5&39.7&43.4&48.1&52.8& 35.0 & 38.3 & 52.9 & 57.7 & 60.8 \\
    DCF  \cite{dcf} & 45.8&59.1&62.1&66.8&68.0&31.8&41.7&46.6&50.3&53.7&39.6&52.1&56.3&60.3&63.3\\
    \rowcolor{Gray}
    Ours   & {\bf 60.1}& {\bf 68.1} & {\bf 69.4} & {\bf 71.2} & {\bf 71.8} & {\bf 45.1} & {\bf 51.5} & {\bf 54.0} & {\bf 56.7} & {\bf 58.8} & {\bf 50.7} & {\bf 58.4} & {\bf 60.0} & {\bf 62.8} & {\bf 64.5} \\
    \bottomrule

  \end{tabular}
  }
\end{adjustbox}
\label{tab:PascalVOC1}
\end{table*}

{\bf MS-COCO.}
Our approach has significantly outperformed previous state-of-the-art methods in few-shot object detection on the MS-COCO dataset. Table \ref{tab:MS-COCO} shows the AP (IOU=50:95) and $\text{AP50}$ of novel classes for two protocols one is gFSOD(w $\slash$ G) and the other is FSOD(wo $\slash$ G).
We report the average results of 10 seeds.
Our results consistently exceed baseline, DeFRCN \cite{defrcn}, across all shots on gFSOD and FSOD. 
Our approach has achieved higher average performance than the existing state-of-the-art in all few-shot settings, with a particularly significant performance increase of around 3\% in fewer-shot scenarios.

{\bf Pascal VOC.}
Our approach consistently outperformed the state-of-the-art in all settings, similar to the MS-COCO dataset. We improved the performance of the baseline and achieved significant performance gains of over 10 points in extremely few scenarios. Table \ref{tab:PascalVOC1} shows the AP50 performance on three different splits of Pascal VOC, with the clean seed results displayed above and the random seed results below. It is worth noting that when comparing the performance difference between DeFRCN \cite{defrcn} and our approach in the 1-shot setting, DeFRCN experiences a significant performance drop in random runs with many missing annotations, whereas our approach maintains relatively stable performance even with a high number of missing annotations.\\
(DeFRCN : 57.0\% $\rightarrow$ 40.2\% vs Ours : 67.2\% $\rightarrow$ 60.1\%).


\subsection{Ablation studies} 
\subsubsection{Effectiveness of CM-CLIP and BNRL}
\begin{table}[ht]
\caption{Effectiveness of different methods in RISF. 
    we report AP of novel classes on MS-COCO under average over multiple runs.
    }
    \centering
    \begin{tabular}{c|c|c|cc|cc|cc}
        \hline
        \multirow{2}{*}{Method} & \multirow{2}{*}{CM-CLIP} & \multirow{2}{*}{BNRL}  &  \multicolumn{2}{c|}{K = 1} &  \multicolumn{2}{c|}{3} &  \multicolumn{2}{c}{10} \\
         & & &AP&AP50&AP&AP50&AP&AP50\\
        \hline
         \multirow{4}{*}{RISF} &  \XSolidBrush & \XSolidBrush  & 5.6&10.3&10.9&19.7&17.2&30.7 \\
          &  \XSolidBrush & \Checkmark  &6.4&13.0& 11.8&22.3&17.4&31.9 \\
          &  \Checkmark & \XSolidBrush  &7.9&15.0&14.2&26.0&19.3&34.3 \\
          &  \Checkmark & \Checkmark  &{\bf 9.2}&{\bf 18.6}&{\bf 15.7}&{\bf 29.6}&{\bf 20.5}&{\bf 37.4} \\
        \bottomrule
    \end{tabular}
    \label{tab:eff_diff_app}
\end{table}

We analyze the effects of BNRL and CM-CLIP on performance in relative ablation under the 1-shot, 3-shot, and 10-shot settings of MS-COCO.

Table \ref{tab:eff_diff_app} provides detailed effectiveness of CM-CLIP and BNRL, showing the AP and AP50 of novel classes.
As mentioned earlier, the random seed configuration represents datasets with severe missing annotations. We take a detailed explanation of RISF as follows:
\begin{enumerate}
    \item As shown by the performance row 1, when CM-CLIP and BNRL are not used (used cross-entropy loss as classification), the detector is the same as DeFRCN \cite{defrcn} not used Prototypical Calibration Block (PCB) and has the lowest performance in all cases except the AP in 10-shot.
    \item Row 2 demonstrates that when BNRL is used, overall performance increases (ranging from 0.2$\%$ to 2.7$\%$).
    \item In row 3, CM-CLIP improves the AP performance in consistently all scenarios (ranging from 2.1$\%$ to 4.7$\%$),
    besides, in the case of 1-shot, AP performance is enormously increased from 5.6\% to 7.9\% and AP50 is increased from 10.3\% to 15.0\%. 
    Therefore, CM-CLIP could be widely applicable to increase performance regardless of the situation.
    \item Row 4 represents the performance when both CM-CLIP and BNRL are used together. 
    The combination of CM-CLIP and BNRL yields the best performance in all scenarios. 
    Notice the performance difference in using BNRL (rows 1-2 and rows 3-4). The performance increase is much greater when using CM-CLIP and BNRL together than when using BNRL alone.
    This experimental result means that the CM-CLIP and BNRL are mutually beneficial.
\end{enumerate}

For that reason, CM-CLIP and BNRL seem to have functionalities that complement each other's drawbacks for a particular case. The drawback of BNRL is its lower concentration for background, which allows the detector to predict high foreground classification scores for actual background instances. This drawback can be compensated for by the classification power of CM-CLIP. On the other hand, though CM-CLIP re-scores the classification score, does not change the classes predicted by the detector. Thus the CM-CLIP is vulnerable to hard negatives, and BNRL's mirror term can complement this weakness. 

\subsubsection{Analysis of the CM-CLIP}

\begin{figure}[h]
    \centering
    \begin{subfigure}[]{0.48\textwidth}
    \includegraphics[width=\linewidth]{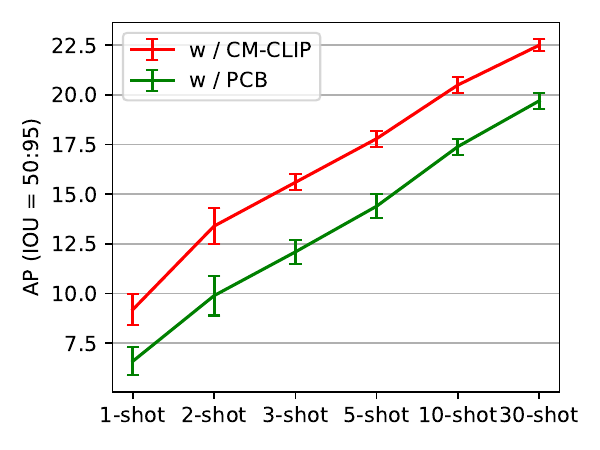}
    \caption{MS-COCO}
    \end{subfigure}
    \begin{subfigure}[]{0.48\textwidth}
    \includegraphics[width=\linewidth]{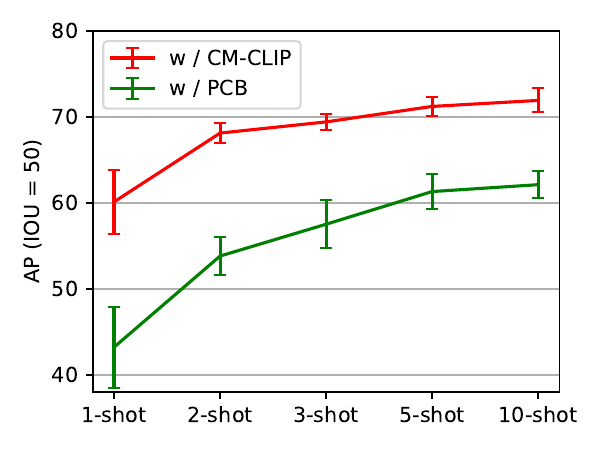}
    \caption{Pascal VOC split 1}
    \end{subfigure}
    \caption{The effectiveness of CM-CLIP on various k-shot settings.
    The average and 95$\%$ confidence interval are computed on 10 random seeds for each shot.}
    \label{fig:cm-clip-result}
\end{figure}

We explore the influence of CM-CLIP for each k-shot setting. 
Figure \ref{fig:cm-clip-result} (a) and (b) show comparisons with CM-CLIP and PCB in RISF. 
Our experimental results showed that using CM-CLIP in all settings was beneficial, and even in Figure \ref{fig:cm-clip-result} (b), the performance with CM-CLIP in the 2-shot setting significantly outperformed the performance with PCB in the 10-shot setting. 
The main difference between PCB and CM-CLIP is that the former calibrates using prototype vectors from images, and the latter calibrates using text embedding from language information.

\begin{figure}[h]
    \centering
    \includegraphics[width=0.45\linewidth]{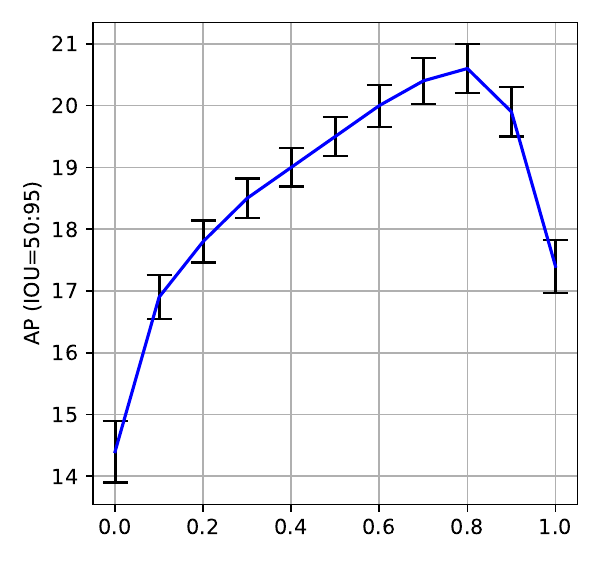}
    \caption{The effectiveness of hyperparameter $\mathbf{c}$ in CM-CLIP. We report the average performance and 95\% confidence interval results for 10 seeds on the MS-COCO 10shot setting.}
\label{fig:cmclip}
\end{figure}

Figure \ref{fig:cmclip} illustrates the performance of CM-CLIP with (a) respect to the parameter $\mathbf{c}$ and (b) the influence of different prompts.
As shown in Equation (2) of the main paper, the final classification score $\mathbf{S}$ depends on the value of $\mathbf{c}$, where a smaller value results in a stronger reliance on the score predicted by CM-CLIP, and a larger value reduces its influence.
In Figure \ref{fig:cmclip} (a), the optimal performance is achieved when $\mathbf{c}$ is appropriately 0.8.

\begin{figure}[h]
    \centering
    \includegraphics[width=0.5\linewidth]{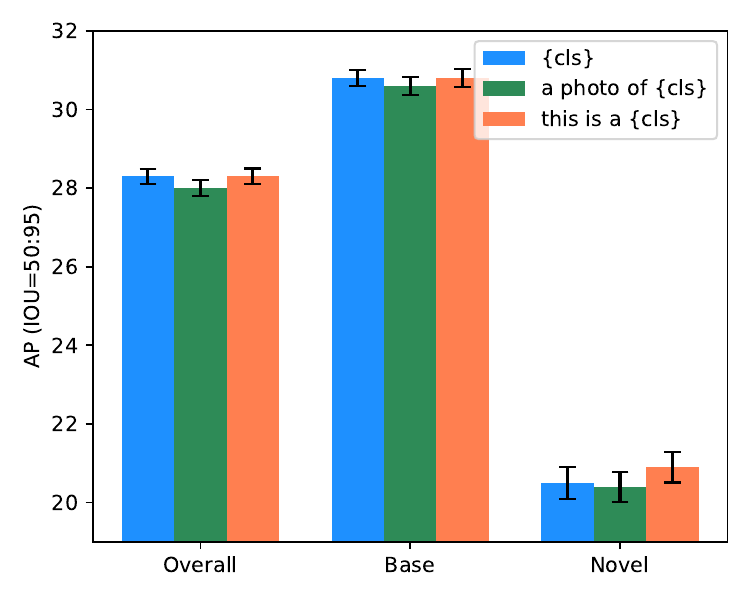}
    \caption{The effectiveness of prompt configurations in CM-CLIP. We report the average performance and 95\% confidence interval results for 10 seeds on the MS-COCO 10shot setting.}
    \label{fig:enter-label}
\end{figure}

Recently, NLP area \cite{gpt, cot} shows that the performance of language model can be increased by applying prompting.
Figure \ref{fig:enter-label} shows that the composition of prompt set input to the CM-CLIP can impact performance.
Specifically, using ``this is a $\{\text{class name}\}$'' as the text input to CM-CLIP leads to an improvement of 0.4$\%$ in novel AP compared to using ``$\{\text{class name}\}$'' alone.

\begin{table}[h]
\caption{Comparison of the vision encoder of CM-CLIP on MS-COCO 10shot dataset. 
The superscript * indicates that CM-CLIP skips re-scoring for base classes. Throughput is measured using an A100 GPU.}
\centering
\begin{adjustbox}{width=1.0\textwidth}
{
\begin{tabular}{l|cc|cc|cc|c|c|c}
     \toprule
     \multirow{2}{*}{Vision Encoder}& \multicolumn{2}{c|}{overall \#80} &\multicolumn{2}{c|}{base \#60}&\multicolumn{2}{c|}{novel\#20}& \multirow{1}{*}{\# Paramters} & \multirow{1}{*}{FLOPs }& {thoughput} \\
      & AP & AP$_{50}$ & AP & AP$_{50}$ & AP & AP$_{50}$ & (Detector/ CM-CLIP) & (Detector/ CM-CLIP) & (image/s) \\
     \hline
     None &28.1&45.3&31.7&49.7&17.4&31.9&  53M &343G &25.6 \\
     ViT-B/16 &28.3 {\scriptsize \color{red}+0.2}&45.6 {\scriptsize \color{red}+0.3}&30.8 {\scriptsize \color{blue}-0.9} &48.0 {\scriptsize \color{blue}-1.7}&20.9 {\scriptsize \color{red}+3.5}&38.3 {\scriptsize \color{red}+6.4}&  53M/ 86M &343G/ 18G&15.7  \\
     ViT-L/14@336px &28.3 {\scriptsize \color{red}+0.2}&45.3&30.8 {\scriptsize \color{blue}-0.9}&48.0 {\scriptsize \color{blue}-1.7}&20.5 {\scriptsize \color{red}+3.1}&37.4 {\scriptsize \color{red}+5.5}&  53M/ 304M &343G/ 191G&4.2\\
     \hline
     ViT-B/16$^*$ &28.8 {\scriptsize \color{red}+0.7}&46.6 {\scriptsize \color{red}+1.3}&31.6 {\scriptsize \color{blue}-0.1} &49.7&20.3 {\scriptsize \color{red}+2.9}&37.2 {\scriptsize \color{red}+5.3}&  53M/ 86M &343G/ 18G&17.8  \\
     ViT-L/14@336px$^*$ &29.0 {\scriptsize \color{red}+0.9}&46.9{\scriptsize \color{red}+1.6}&31.7 &49.7&21.0 {\scriptsize \color{red}+3.6}&38.4 {\scriptsize \color{red}+6.5}&  53M/ 304M &343G/ 191G&6.9\\
     \bottomrule
\end{tabular}
}
\end{adjustbox}
\label{tab:cmclip}
\end{table}

Intuitively, CM-CLIP extracts image features and thus demands additional computation costs.
Table \ref{tab:cmclip} {\bf UP} indicates the comparison between using ViT-base or ViT-Large as the vision encoder in CM-CLIP, and the scenario where CM-CLIP is not utilized at all.
In this case, the computation cost increases in proportion to the number of predicted instances. 
Consequently, in dense object scenarios such as MS-COCO, there is an inference latency of approximately 40\% when CM-CLIP (ViT-base) is applied.
Furthermore, though CM-CLIP exhibits significant performance improvement in novel classes, there is a slight decrease in performance in base classes.
To address this issue, it is possible to allow CM-CLIP to skip re-scoring for base classes, resulting in both performance increase and reduced computation.
Then, as shown in Table \ref{tab:cmclip} {\bf DOWN}, the degradation of performance for base classes is nearly zero, and inference latency is reduced by approximately 10\% for ViT-base.
Moreover, ViT-base is still powerful in terms of increasing performance, even though it is faster than ViT-Large by around x3.7.

\subsubsection{Analysis of the BNRL}

\begin{figure}[h]
\centering
\includegraphics[width=0.6\linewidth]{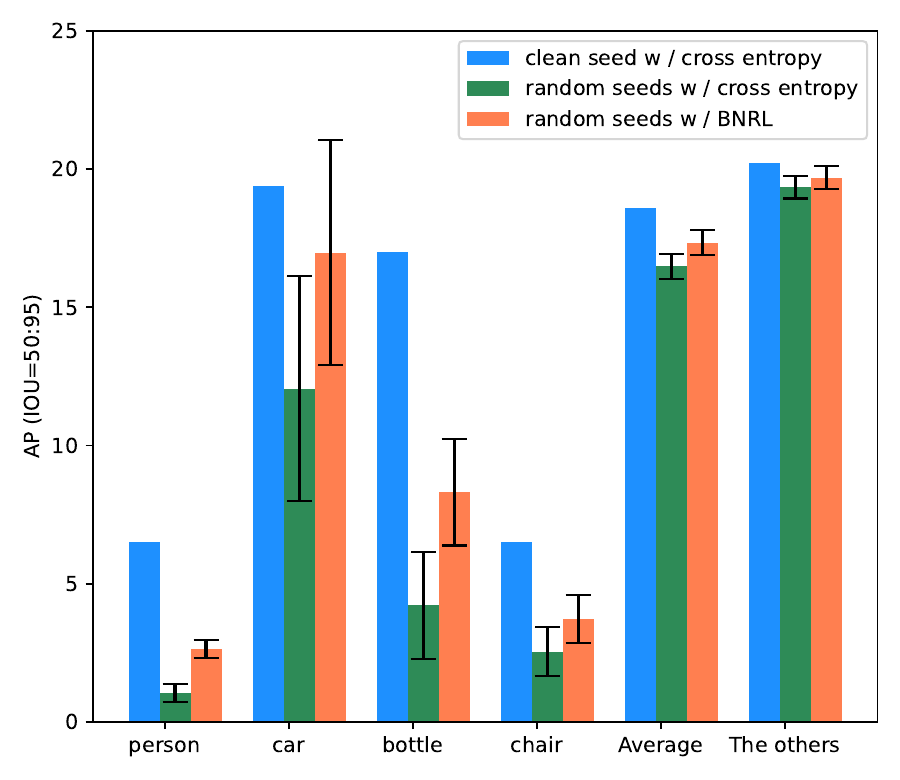} 
\caption{Illustration of the performance comparison between BNRL and cross-entropy loss. }
\label{fig:wrapfig}
\end{figure}
As mentioned in Section 3.3, BNRL can alleviate performance decrease caused by missing annotation.
Figure \ref{fig:wrapfig} shows the effectiveness of BNRL in FSOD setting.
The blue bar indicates the detector trained on clean seed and thus achieves the highest scores in each class. 
The green and orange bars indicate the detectors trained on random seeds and thus their AP performance is lower than the blue bar.
However, the orange bar is higher than the green bar so the orange bar uses BNRL though the green bar uses cross entropy loss.
For the 4 classes (person, car, bottle, chair) with the highest number of missing annotations in MS-COCO, it is more effective in increasing performance to use BNRL than the remaining 16 classes (the others) with the low number of missing annotations.

\begin{figure}[hbt!]
    \centering

    \begin{subfigure}[]{0.32\textwidth}
    \includegraphics[width=\linewidth]{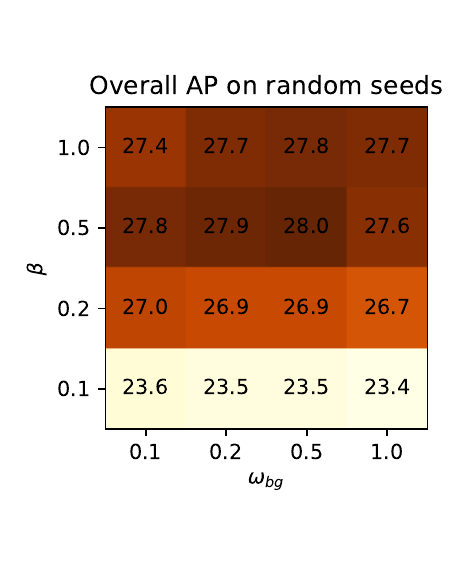}
    \end{subfigure}
    \begin{subfigure}[]{0.32\textwidth}
    \includegraphics[width=\linewidth]{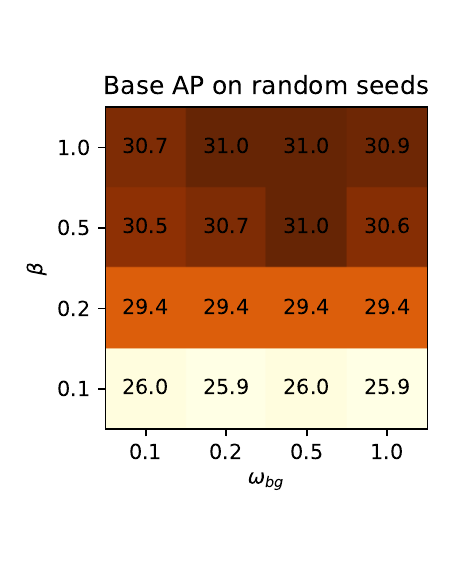}
    \end{subfigure}
    \begin{subfigure}[]{0.32\textwidth}
    \includegraphics[width=\linewidth]{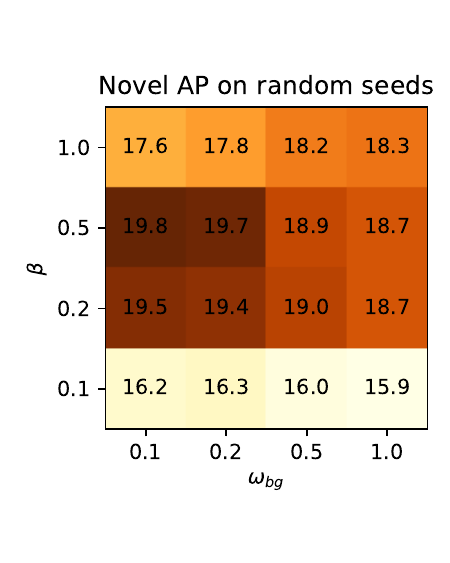}
    \end{subfigure}

    \caption{The effectiveness of hyperparameters in BNRL. 
    The left, middle, and right of the figure represent the AP for overall, base, and novel classes, respectively.
    The vertical and horizontal axes represent the $\beta$ and $\omega_{bg}$ in BNRL.
    }
    \label{fig:BNRL}
\end{figure}

We carefully explore the effectiveness of the hyperparameters in BNRL on MS-COCO 10-shot in Figure \ref{fig:BNRL}. 
In Section 3.3, a smaller value of $\beta$ strengthens the learning of hard negative samples, while a smaller value of $\omega_{bg}$ leads to the ignoring of the background.
If $\beta$ and $\omega_{bg}$ are 1.0, they are equal to focal loss.
The hyperparameters in BNRL have varying effects on performance depending on the number of missing annotations. 
For random seeds with a high number of missing annotations (Figure \ref{fig:BNRL}), when $\alpha$ and $\omega_{bg}$ are set to moderate, the optimal performance is achieved.
This result can be interpreted as $\beta$ and $\omega_{bg}$ are complementary to each other.
For example, as $\beta$ grows, it becomes robust to hard negative, but it becomes more vulnerable to incorrect labels because high confidence predictions can be hard negative for mislabeled objects.  

In FSOD, most of the incorrect labels are missing annotations, and thus lowering $\omega_{bg}$ can alleviate the problem of being vulnerable to incorrect labels.
Therefore, changing only one $\beta$ or $\omega_{bg}$ has a poor effect on performance, but changing both hyperparameters at the same time has a good effect on performance, as shown in Figure \ref{fig:BNRL}.

\subsubsection{Different backbone networks}

In the transfer learning paradigm, most research is based on the Faster-RCNN architecture with a ResNet \cite{resnet} backbone network. In this section, we present performance results based on different backbone network sizes.
Table \ref{tab:backbone} compares the SWIN transformer \cite{swin} backbone with the ResNet-101 used in the main paper. When using SWIN-Tiny, there is a decrease in performance, while using SWIN-Large leads to a significant performance improvement. Thus, the general rule in deep learning that larger models tend to perform better holds true in FSOD as well.

\begin{table}[h]
\caption{Comparison of different backbone networks of the detector on MS-COCO dataset.}
\centering
\begin{adjustbox}{width=1.0\textwidth}
{
\begin{tabular}{c|l|ccc|ccc|ccc}
     \toprule
     \multirow{2}{*}{Method}&\multirow{2}{*}{Backbone}& \multicolumn{3}{c|}{5-shot} &\multicolumn{3}{c|}{10-shot}&\multicolumn{3}{c}{30-shot}  \\
      && AP & bAP & nAP & AP & bAP & nAP & AP & bAP & nAP \\
     \hline
     \multirow{3}{*}{RISF} &ResNet101\cite{resnet}& 26.6 & 29.3& 18.4 & 28.7& 31.3 & 21.1 & 29.8 & 31.9 & 23.6 \\
     &SWIN-Tiny\cite{swin} &21.6 {\scriptsize \color{blue}-5.0}&23.6 {\scriptsize \color{blue}-3.0}&15.7 {\scriptsize \color{blue}-2.7}&24.1 {\scriptsize \color{blue}-4.6}&26.0 {\scriptsize \color{blue}-5.3}&18.5 {\scriptsize \color{blue}-2.6}&26.7 {\scriptsize \color{blue}-3.1}&27.9 {\scriptsize \color{blue}-4.0}&23.2 {\scriptsize \color{blue}-0.4}\\
     &SWIN-Large\cite{swin} & 32.3 {\scriptsize \color{red}+5.7}&35.7 {\scriptsize \color{red}+6.4}&22.3 {\scriptsize \color{red}+3.9}&34.8 {\scriptsize \color{red}+6.1}&37.9 {\scriptsize \color{red}+6.6}&25.5 {\scriptsize \color{red}+4.4}&37.5 {\scriptsize \color{red}+6.7}&39.4 {\scriptsize \color{red}+7.5}&31.9 {\scriptsize \color{red}+8.3}  \\
     \midrule
     \multicolumn{10}{l}{\textit{Results averaged over multiple random runs:}}\\
     \multirow{3}{*}{RISF} &ResNet101\cite{resnet}&25.7&28.3&17.8&28.2&30.8&20.5&29.5&31.8&22.5 \\
     &SWIN-Tiny\cite{swin} & 20.4 {\scriptsize \color{blue}-5.3}&22.7 {\scriptsize \color{blue}-5.6}&13.6 {\scriptsize \color{blue}-4.2}&23.5 {\scriptsize \color{blue}-4.7}&25.6 {\scriptsize \color{blue}-5.2}&17.2 {\scriptsize \color{blue}-3.3}&27.0 {\scriptsize \color{blue}-2.5}&28.6 {\scriptsize \color{blue}-3.2}&22.1 {\scriptsize \color{blue}-0.4} \\
     &SWIN-Large\cite{swin}&29.7 {\scriptsize \color{red}+4.0}&32.9 {\scriptsize \color{red}+4.6}&20.1 {\scriptsize \color{red}+2.3}&33.1 {\scriptsize \color{red}+4.9}&35.9 {\scriptsize \color{red}+5.1}&24.8 {\scriptsize \color{red}+4.3}&37.1 {\scriptsize \color{red}+7.6}&39.2 {\scriptsize \color{red}+7.4}&30.9 {\scriptsize \color{red}+8.4}\\
     \bottomrule
\end{tabular}
}
\end{adjustbox}
\label{tab:backbone}
\end{table}

\subsection{Complete Results of gFSOD}

We report the average AP with 95\% confidence interval of the RISF using 10 different seeds on Pascal VOC and MS-COCO in Table \ref{tab:gfsod} and Table \ref{tab:gfsod_coco}.
As shown in Table \ref{tab:gfsod} RISF outperforms DeFRCN and TFA on all gFSOD settings of Pascal VOC.
In Table \ref{tab:gfsod_coco}, sometimes, RISF loses to DeFRCN or TFA, but wins in most cases of MS-COCO.
In these experiments, the output layers of the base model are not randomly initialized when starting fine-tuning, similar to DeFRCN and TFA, to ensure a fair evaluation of gFSOD.

\begin{table}[hbt!]
    \caption{gFSOD experimental results on MS-COCO.}
    \centering
    \begin{tabular}{c|c|ccc|c|c}
        \toprule
         \multirow{2}{*}{\# shot} & \multirow{2}{*}{Method} & \multicolumn{3}{c|}{Overall \#80} &Base \#60 & Novel \#20  \\
         && AP & AP50 &AP75 & AP&AP \\
         \hline
         \multirow{4}{*}{1} & FRCN+ft \cite{metarcnn} & 16.2$\pm$0.9& 25.8$\pm$1.2 &17.6$\pm$1.0 & 21.0$\pm$1.2 & 1.7$\pm$0.2 \\
         & TFA \cite{tfa} & 24.4$\pm$0.6 & 39.8$\pm$0.8 & 26.1$\pm$0.8 & 31.9$\pm$0.7 & 1.9$\pm$0.4 \\
         & DeFRCN \cite{defrcn} & 24.0$\pm$0.4 & 36.9$\pm$0.6 & 26.2$\pm$0.4 & 30.4$\pm$0.4 & 4.8$\pm$0.6 \\
         & \cellcolor{gray!30} OURS & \cellcolor{gray!30} 25.0 $\pm$ 0.3  {\scriptsize \color{red}+0.6}& \cellcolor{gray!30}39.7 $\pm$ 0.6  {\scriptsize \color{blue}-0.1}&\cellcolor{gray!30} 26.8 $\pm$0.4 {\scriptsize \color{red}+0.7}& \cellcolor{gray!30}30.6 $\pm$0.4 {\scriptsize \color{blue}-1.3}&\cellcolor{gray!30} 8.2 $\pm$ 0.9 {\scriptsize \color{red}+3.4}\\
         \hline
         \multirow{4}{*}{2} & FRCN+ft \cite{metarcnn} & 15.8$\pm$0.7& 25.0$\pm$1.1 &17.3$\pm$0.7 & 20.0$\pm$0.9 & 3.1$\pm$0.3 \\
         & TFA \cite{tfa} & 24.9$\pm$0.6 & 40.1$\pm$0.9 & 27.0$\pm$0.7 & 31.9$\pm$0.7 & 3.9$\pm$0.4 \\
         & DeFRCN \cite{defrcn}& 25.7$\pm$0.5 & 39.6$\pm$0.8 & 28.0$\pm$0.5 & 31.4$\pm$0.4 & 8.5$\pm$0.8 \\
         & \cellcolor{gray!30}OURS & \cellcolor{gray!30}26.7 $\pm$ 0.5 {\scriptsize \color{red}+1.0}& \cellcolor{gray!30}42.6 $\pm$ 0.8 {\scriptsize \color{red}+2.5}& \cellcolor{gray!30}28.6 $\pm$ 0.6 {\scriptsize \color{red}+0.6}& \cellcolor{gray!30}31.4 $\pm$ 0.4 {\scriptsize \color{blue}-0.5}& \cellcolor{gray!30}12.7 $\pm$ 0.9 {\scriptsize \color{red}+4.2}\\
         \hline
         \multirow{4}{*}{3} & FRCN+ft \cite{metarcnn} & 15.0$\pm$0.7& 23.9$\pm$1.2 &16.4$\pm$0.7 & 18.8$\pm$0.9 & 3.7$\pm$0.4 \\
         & TFA \cite{tfa} & 25.3$\pm$0.6 & 40.4$\pm$1.0 & 27.6$\pm$0.7 & 32.0$\pm$0.7 & 5.1$\pm$0.6 \\
         & DeFRCN \cite{defrcn}& 26.6$\pm$0.4 & 41.1$\pm$0.7 & 28.9$\pm$0.4 & 32.1$\pm$0.3 & 10.7$\pm$0.8 \\
         & \cellcolor{gray!30} OURS & \cellcolor{gray!30} 27.8 $\pm$ 0.2 {\scriptsize \color{red}+1.2} & \cellcolor{gray!30} 44.1 $\pm$0.5 {\scriptsize \color{red}+3.0}& \cellcolor{gray!30} 29.8 $\pm$0.2 {\scriptsize \color{red}+0.9}& \cellcolor{gray!30}32.2 $\pm$0.2 {\scriptsize \color{red}+0.1}&\cellcolor{gray!30} 14.6 $\pm$0.6 {\scriptsize \color{red}+3.9}\\
         \hline
         \multirow{4}{*}{5} & FRCN+ft \cite{metarcnn} & 14.4$\pm$0.8& 23.0$\pm$1.3 &15.6$\pm$0.8 & 17.6$\pm$0.9 & 4.6$\pm$0.5 \\
         & TFA \cite{tfa} & 25.9$\pm$0.6 & 41.2$\pm$0.9 & 28.4$\pm$0.6 & 32.3$\pm$0.6 & 7.0$\pm$0.7 \\
         & DeFRCN \cite{defrcn}& 27.8$\pm$0.3 & 43.0$\pm$0.6 & 30.2$\pm$0.3 & 32.6$\pm$0.3 & 13.6$\pm$0.7 \\
         & \cellcolor{gray!30} OURS & \cellcolor{gray!30} 28.8 $\pm$0.2 {\scriptsize \color{red}+1.0}& \cellcolor{gray!30} 45.8 $\pm$0.4 {\scriptsize \color{red}+2.8}& \cellcolor{gray!30} 30.7 $\pm$0.2 {\scriptsize \color{red}+0.5}& \cellcolor{gray!30} 32.9 $\pm$0.2 {\scriptsize \color{red}+0.3}& \cellcolor{gray!30} 16.6$\pm$0.4 {\scriptsize \color{red}+3.0}\\
         \hline
         \multirow{4}{*}{10} & FRCN+ft \cite{metarcnn} & 13.4$\pm$0.8& 21.8$\pm$1.7 &14.5$\pm$0.9 & 16.1$\pm$1.0 & 5.5$\pm$0.9 \\
         & TFA \cite{tfa} & 26.6$\pm$0.5 & 42.2$\pm$0.8 & 29.0$\pm$0.6 & 32.4$\pm$0.6 & 9.1$\pm$0.5 \\
         & DeFRCN \cite{defrcn}& 29.7$\pm$0.2 & 46.0$\pm$0.5 & 32.1$\pm$0.2 & 34.0$\pm$0.2 & 16.8$\pm$0.6 \\
         & \cellcolor{gray!30} OURS & \cellcolor{gray!30} 30.2 $\pm$0.3 {\scriptsize \color{red}+0.5}& \cellcolor{gray!30} 48.1$\pm$0.5 {\scriptsize \color{red}+2.1}& \cellcolor{gray!30} 32.2$\pm$0.3 {\scriptsize \color{red}+0.1}& \cellcolor{gray!30} 33.9$\pm$0.2 {\scriptsize \color{blue}-0.1}& \cellcolor{gray!30} 19.1$\pm$0.6 {\scriptsize \color{red}+2.3}\\
         \hline
         \multirow{4}{*}{30} & FRCN+ft \cite{metarcnn} & 13.5$\pm$1.0& 21.8$\pm$1.9 &14.5$\pm$1.0 & 15.6$\pm$1.0 & 7.4$\pm$1.1 \\
         & TFA \cite{tfa} & 28.7$\pm$0.4 & 44.7$\pm$0.9 & 31.5$\pm$0.4 & 34.2$\pm$0.4 & 12.1$\pm$0.4 \\
         & DeFRCN \cite{defrcn}& 31.4$\pm$0.1 & 48.8$\pm$0.2 & 33.9$\pm$0.1 & 34.8$\pm$0.1 & 21.2$\pm$0.4 \\
         & \cellcolor{gray!30} OURS & \cellcolor{gray!30} 31.9$\pm$0.1 {\scriptsize \color{red}+0.5}& \cellcolor{gray!30} 50.3$\pm$0.3 {\scriptsize \color{red}+1.5}& \cellcolor{gray!30} 34.1$\pm$0.2 {\scriptsize \color{red}+0.2}& \cellcolor{gray!30} 35.1$\pm$0.1 {\scriptsize \color{red}+0.3}& \cellcolor{gray!30} 22.4$\pm$0.5{\scriptsize \color{red}+1.2} \\

         \bottomrule
    \end{tabular}
    
    \label{tab:gfsod_coco}
\end{table}
\newpage

\begin{table}[hbt!]
    \caption{gFSOD experimental results for 1,2,3,5, and 10-shot settings on Pascal VOC.}

    \centering
    \begin{adjustbox}{width=0.79\textwidth}
        {
    \begin{tabular}{c|c|c|ccc|c|cc}
        \toprule
         \multirow{2}{*}{Split}&\multirow{2}{*}{\#shot}&\multirow{2}{*}{Method}&\multicolumn{3}{c|}{Overall \#20} & \multicolumn{1}{c|}{Base \#20} & \multicolumn{2}{c}{Novel \#5}  \\
         &&&AP&AP50&AP75&AP&AP&AP50 \\
         \hline
         \multirow{25}{*}{ Set 1}&\multirow{5}{*}{1}&FSRW \cite{metayolo} &27.6$\pm$0.5&50.8$\pm$0.9&26.5$\pm$0.6&34.1$\pm$0.5&8.0$\pm$1.0&14.2  \\
         &&FRCN-ft \cite{metarcnn}&30.2$\pm$0.6&49.4$\pm$0.7&32.2$\pm$0.9&38.2$\pm$0.8&6.0$\pm$0.7&9.9\\
         &&TFA \cite{tfa}&40.6$\pm$0.5&64.5$\pm$0.6&44.7$\pm$0.6&49.4$\pm$0.4&14.2$\pm$1.4&25.3\\
         && DeFRCN \cite{defrcn}&42.0$\pm$0.6&66.7$\pm$0.8&45.5$\pm$0.7&48.4$\pm$0.4&22.5$\pm$1.7&40.2\\
         && \cellcolor{gray!30}Ours&\cellcolor{gray!30}44.1$\pm$0.9 {\scriptsize \color{red}+2.1}&\cellcolor{gray!30}70.5$\pm$1.6 {\scriptsize \color{red}+3.8}&\cellcolor{gray!30}47.5$\pm$1.0 {\scriptsize \color{red}+2.0}&\cellcolor{gray!30} 49.4$\pm$0.8 {\scriptsize \color{red}+1.0}&\cellcolor{gray!30} 28.1$\pm$2.4 {\scriptsize \color{red}+5.6}&\cellcolor{gray!30} 51.0$\pm$4.5 {\scriptsize \color{red}+10.8}\\
         \cmidrule{2-9}
         &\multirow{5}{*}{2}&FSRW \cite{metayolo} & 28.7$\pm$0.4 & 52.5$\pm$0.6& 27.7$\pm$0.5& 33.9 $\pm$0.5&13.2$\pm$1.0&23.6 \\
         &&FRCN-ft \cite{metarcnn}&30.5$\pm$0.6&49.4$\pm$0.8&32.6$\pm$0.7&37.3$\pm$0.7&9.9$\pm$0.9&15.6\\
         &&TFA \cite{tfa}&42.6$\pm$0.3&67.1$\pm$0.4&47.0$\pm$0.4&49.6$\pm$0.3&21.7$\pm$1.0&36.4\\
         &&DeFRCN \cite{defrcn}&44.3$\pm$ 0.4&70.2$\pm$0.5&48.0$\pm$0.6&49.1$\pm$0.3&30.6$\pm$1.2&53.6\\
         && \cellcolor{gray!30}Ours&\cellcolor{gray!30}46.6$\pm$0.6 {\scriptsize \color{red}+2.3}&\cellcolor{gray!30}74.4$\pm$0.7 {\scriptsize \color{red}+4.2}&\cellcolor{gray!30}50.1$\pm$0.8 {\scriptsize \color{red}+2.1}&\cellcolor{gray!30}49.8$\pm$0.5 {\scriptsize \color{red}+0.7}&\cellcolor{gray!30}37.0$\pm$1.4 {\scriptsize \color{red}+6.4}&\cellcolor{gray!30}65.1$\pm$2.1 {\scriptsize \color{red}+11.5}\\
         \cmidrule{2-9}
         &\multirow{5}{*}{3}&FSRW \cite{metayolo} & 29.5$\pm$0.3&53.3$\pm$0.6&28.6$\pm$0.4&33.8$\pm$0.3&16.8$\pm$0.9&29.8 \\
         &&FRCN-ft \cite{metarcnn}&31.8$\pm$0.5&51.4$\pm$0.8&34.2$\pm$0.6&37.9$\pm$0.5&13.7$\pm$1.0&21.6\\
         &&TFA \cite{tfa}&43.7$\pm$0.3&68.5$\pm$0.4&48.3$\pm$0.4&49.8$\pm$0.3&25.4$\pm$0.9&42.1\\
         &&DeFRCN \cite{defrcn}&45.3$\pm$0.3&71.5$\pm$0.4&49.0$\pm$0.5&49.3$\pm$0.3&33.7$\pm$0.8&58.2\\
         && \cellcolor{gray!30}Ours&\cellcolor{gray!30}47.4$\pm$0.5 {\scriptsize \color{red}+1.9}&\cellcolor{gray!30}75.2$\pm$0.6 {\scriptsize \color{red}+3.7}&\cellcolor{gray!30}51.1$\pm$0.9 {\scriptsize \color{red}+2.1}&\cellcolor{gray!30}50.1$\pm$0.4 {\scriptsize \color{red}+0.8}&\cellcolor{gray!30}39.2$\pm$1.5 {\scriptsize \color{red}+5.5}&\cellcolor{gray!30}67.9$\pm$1.8 {\scriptsize \color{red}+9.7}\\
         \cmidrule{2-9}
         &\multirow{5}{*}{5}&FSRW \cite{metayolo} & 30.4$\pm$0.3&54.6$\pm$0.5&29.6$\pm$0.4&33.7$\pm$0.3&20.6$\pm$0.8&36.5 \\
         &&FRCN-ft \cite{metarcnn}&32.7$\pm$0.5&52.5$\pm$0.8&35.0$\pm$0.6&37.6$\pm$0.4&17.9$\pm$1.1&28.0\\
         &&TFA \cite{tfa}&44.8$\pm$0.3&70.1$\pm$0.4&49.4$\pm$0.4&50.1$\pm$0.2&28.9$\pm$0.8&47.9\\
         &&DeFRCN \cite{defrcn}&46.4$\pm$0.3&73.1$\pm$0.3&50.4$\pm$0.4&49.6$\pm$0.3&37.3$\pm$0.8&63.6\\
         && \cellcolor{gray!30}Ours&\cellcolor{gray!30}48.5$\pm$0.4 {\scriptsize \color{red}+2.1}&\cellcolor{gray!30}76.3$\pm$0.4 {\scriptsize \color{red}+3.2}&\cellcolor{gray!30}52.8$\pm$0.8 {\scriptsize \color{red}+2.4}&\cellcolor{gray!30}50.6$\pm$0.4 {\scriptsize \color{red}+1.0}&\cellcolor{gray!30}42.3$\pm$0.8 {\scriptsize \color{red}+5.0}&\cellcolor{gray!30}72.0$\pm$0.8 {\scriptsize \color{red}+8.4}\\
         \cmidrule{2-9}
         &\multirow{5}{*}{10}& FRCN-ft \cite{metarcnn}&33.3$\pm$0.4&53.8$\pm$0.6&35.5$\pm$0.4&36.8$\pm$0.4&22.7$\pm$0.9&52.0 \\
         &&TFA \cite{tfa}&45.8$\pm$0.2&71.3$\pm$0.3&50.4$\pm$0.3&50.4$\pm$0.2&32.0$\pm$0.6&52.8\\
         &&DeFRCN \cite{defrcn}&47.2$\pm$0.2&74.0$\pm$0.3&51.3$\pm$0.3&49.9$\pm$0.2&39.8$\pm$0.7&66.5\\
         && \cellcolor{gray!30}Ours&\cellcolor{gray!30}48.8$\pm$0.4 {\scriptsize \color{red}+1.6}&\cellcolor{gray!30}76.5$\pm$0.5 {\scriptsize \color{red}+2.5}&\cellcolor{gray!30}53.0$\pm$0.5 {\scriptsize \color{red}+1.7}&\cellcolor{gray!30}50.7$\pm$0.3 {\scriptsize \color{red}+0.8}&\cellcolor{gray!30}43.1$\pm$1.2 {\scriptsize \color{red}+3.3}&\cellcolor{gray!30}72.0$\pm$2.2 {\scriptsize \color{red}+5.5}\\
         \midrule
         \multirow{25}{*}{ Set 2}&\multirow{5}{*}{1}&FSRW \cite{metayolo} &28.4$\pm$0.5&51.7$\pm$0.9&27.3$\pm$0.6&35.7$\pm$0.5&6.3$\pm$0.9&12.3\\
         &&FRCN-ft \cite{metarcnn} &30.3$\pm$0.5&49.7$\pm$0.5&32.3$\pm$0.7&38.8$\pm$0.6&5.0$\pm$0.6&9.4\\
         &&TFA \cite{tfa}&36.7$\pm$0.6&59.9$\pm$0.8&39.3$\pm$0.8&45.9$\pm$0.7&7.0$\pm$1.2&18.3\\
         &&DeFRCN \cite{defrcn}&40.7$\pm$0.5&64.8$\pm$0.7&43.8$\pm$0.6&49.6$\pm$0.4&14.6$\pm$1.5&29.5\\
         && \cellcolor{gray!30}Ours&\cellcolor{gray!30}42.6$\pm$0.8 {\scriptsize \color{red}+1.9}&\cellcolor{gray!30}68.0$\pm$1.0 {\scriptsize \color{red}+3.6}&\cellcolor{gray!30}45.4$\pm$1.1 {\scriptsize \color{red}+1.6}&\cellcolor{gray!30}50.4$\pm$1.0 {\scriptsize \color{red}+0.8}&\cellcolor{gray!30}19.1$\pm$2.2 {\scriptsize \color{red}+4.5}&\cellcolor{gray!30}38.3$\pm$3.4 {\scriptsize \color{red}+8.8}\\
         \cmidrule{2-9}
         &\multirow{5}{*}{2}&FSRW \cite{metayolo} & 29.4$\pm$0.3&53.1$\pm$0.6&28.5$\pm$0.4&35.8$\pm$0.4&9.9$\pm$0.7&19.6  \\
         &&FRCN-ft \cite{metarcnn}&30.7$\pm$0.5&49.7$\pm$0.7&32.9$\pm$0.6&38.4$\pm$0.5&7.7$\pm$0.8&13.8\\
         &&TFA \cite{tfa}&39.0$\pm$0.4&63.0$\pm$0.5&42.1$\pm$0.6&47.3$\pm$0.4&14.1$\pm$0.9&27.5\\
         &&DeFRCN \cite{defrcn}&42.7$\pm$0.3&67.7$\pm$0.5&45.7$\pm$0.5&50.3$\pm$0.2&20.5$\pm$1.0&39.7\\
         && \cellcolor{gray!30}Ours&\cellcolor{gray!30}44.4$\pm$0.6 {\scriptsize \color{red}+1.7}&\cellcolor{gray!30}70.8$\pm$0.7 {\scriptsize \color{red}+3.1}&\cellcolor{gray!30}47.4$\pm$1.0 {\scriptsize \color{red}+1.7}&\cellcolor{gray!30}50.8$\pm$0.4 {\scriptsize \color{red}+0.5}&\cellcolor{gray!30}25.1$\pm$1.7 {\scriptsize \color{red}+4.6}&\cellcolor{gray!30}48.6$\pm$2.1 {\scriptsize \color{red}+8.9}\\
         \cmidrule{2-9}
         &\multirow{5}{*}{3}&FSRW \cite{metayolo} &29.9$\pm$0.3&53.9$\pm$0.4&29.0$\pm$0.4&35.7$\pm$0.3&12.5$\pm$0.7&25.1  \\
         &&FRCN-ft \cite{metarcnn}&31.1$\pm$0.3&50.1$\pm$0.5&33.2$\pm$0.5&38.1$\pm$0.4&9.8$\pm$0.9&17.4\\
         &&TFA \cite{tfa}&40.1$\pm$0.3&64.5$\pm$0.5&43.3$\pm$0.4&48.1$\pm$0.3&16.0$\pm$0.8&30.9\\
         &&DeFRCN \cite{defrcn}&43.5$\pm$0.3&68.9$\pm$0.4&46.6$\pm$0.4&50.6$\pm$0.3&22.9$\pm$1.0&43.4\\
         && \cellcolor{gray!30}Ours&\cellcolor{gray!30}45.0$\pm$0.6 {\scriptsize \color{red}+1.5}&\cellcolor{gray!30}72.0$\pm$0.6 {\scriptsize \color{red}+3.1}&\cellcolor{gray!30}47.9$\pm$0.8 {\scriptsize \color{red}+1.3}&\cellcolor{gray!30}51.0$\pm$0.5 {\scriptsize \color{red}+0.4}&\cellcolor{gray!30}27.0$\pm$1.1 {\scriptsize \color{red}+4.1}&\cellcolor{gray!30}52.7$\pm$1.5 {\scriptsize \color{red}+9.3}\\
         \cmidrule{2-9}
         &\multirow{5}{*}{5}&FSRW \cite{metayolo} & 30.4$\pm$0.4&54.6$\pm$0.5&29.5$\pm$0.5&35.3$\pm$0.3&15.7$\pm$0.8&31.4  \\
         &&FRCN-ft \cite{metarcnn}&31.5$\pm$0.3&50.8$\pm$0.7&33.6$\pm$0.4&37.9$\pm$0.4&12.4$\pm$0.9&21.9\\
         &&TFA \cite{tfa}&40.9$\pm$0.4&65.7$\pm$0.5&44.1$\pm$0.5&48.6$\pm$0.4&17.8$\pm$0.8&34.1\\
         &&DeFRCN \cite{defrcn}&44.6$\pm$0.3&70.2$\pm$0.5&47.8$\pm$0.4&51.0$\pm$0.2&25.8$\pm$0.9&48.1\\
         && \cellcolor{gray!30}Ours&\cellcolor{gray!30}46.2$\pm$0.4 {\scriptsize \color{red}+1.6}&\cellcolor{gray!30}73.4$\pm$0.5 {\scriptsize \color{red}+3.2}&\cellcolor{gray!30}49.4$\pm$0.8 {\scriptsize \color{red}+1.6}&\cellcolor{gray!30}51.6$\pm$0.5 {\scriptsize \color{red}+0.6}&\cellcolor{gray!30}30.0$\pm$1.1 {\scriptsize \color{red}+4.2}&\cellcolor{gray!30}56.9$\pm$1.5 {\scriptsize \color{red}+8.8}\\
         \cmidrule{2-9}
         &\multirow{5}{*}{10}&FRCN-ft \cite{metarcnn}&32.2$\pm$0.3&52.3$\pm$0.4&34.1$\pm$0.4&37.2$\pm$0.3&17.0$\pm$0.8&39.7\\
         &&TFA \cite{tfa}&42.3$\pm$0.3&37.6$\pm$0.4&45.7$\pm$0.3&49.4$\pm$0.2&20.8$\pm$0.6&39.5\\
         &&DeFRCN \cite{defrcn}&45.6$\pm$0.2&71.5$\pm$0.3&49.0$\pm$0.3&51.3$\pm$0.2&29.3$\pm$0.7&52.8\\
         && \cellcolor{gray!30}Ours&\cellcolor{gray!30}47.0$\pm$0.3 {\scriptsize \color{red}+2.4}&\cellcolor{gray!30}73.9$\pm$0.5 {\scriptsize \color{red}+2.4}&\cellcolor{gray!30}50.4$\pm$0.5 {\scriptsize \color{red}+1.4}&\cellcolor{gray!30}51.8$\pm$0.3 {\scriptsize \color{red}+0.5}&\cellcolor{gray!30}32.5$\pm$0.7 {\scriptsize \color{red}+3.2}&\cellcolor{gray!30}59.7$\pm$1.2 {\scriptsize \color{red}+6.9}\\
         \midrule
         \multirow{25}{*}{ Set 3}&\multirow{5}{*}{1}&FSRW \cite{metayolo} &27.5$\pm$0.6&50.0$\pm$1.0&26.8$\pm$0.7& 34.5$\pm$0.7&6.7$\pm$1.0&12.5 \\
         &&FRCN-ft \cite{metarcnn}&30.8$\pm$0.6&49.8$\pm$0.8&32.9$\pm$0.8&39.6$\pm$0.8&4.5$\pm$0.7&8.1\\
         &&TFA \cite{tfa}&40.1$\pm$0.3&63.5$\pm$0.6&43.6$\pm$0.5&50.2$\pm$0.4&9.6$\pm$1.1&17.9\\
         && DeFRCN \cite{defrcn}&41.6$\pm$0.5&66.0$\pm$0.9&44.9$\pm$0.6&49.4$\pm$0.4&17.9$\pm$1.6&35.0\\
         && \cellcolor{gray!30}Ours&\cellcolor{gray!30}43.5$\pm$1.0 {\scriptsize \color{red}+1.9}&\cellcolor{gray!30}69.7$\pm$1.7 {\scriptsize \color{red}+3.7}&\cellcolor{gray!30}46.8$\pm$1.0 {\scriptsize \color{red}+1.9}&\cellcolor{gray!30}50.2$\pm$0.5 {\scriptsize \color{red}+0.8}&\cellcolor{gray!30}23.5$\pm$3.2 {\scriptsize \color{red}+5.6}&\cellcolor{gray!30}45.2$\pm$5.4 {\scriptsize \color{red}+10.2}\\
         \cmidrule{2-9}
         &\multirow{5}{*}{2}&FSRW \cite{metayolo} & 28.7$\pm$0.4&51.8$\pm$0.7&28.1$\pm$0.5&34.5$\pm$0.4&11.3$\pm$0.7&21.3 \\
         &&FRCN-ft \cite{metarcnn}&31.3$\pm$0.5&50.2$\pm$0.9&33.5$\pm$0.6&39.1$\pm$0.5&8.0$\pm$0.8&13.9\\
         &&TFA \cite{tfa}&41.8$\pm$0.4&65.6$\pm$0.5&45.3$\pm$0.4&50.7$\pm$0.3&15.1$\pm$1.3&27.2\\
         &&DeFRCN \cite{defrcn}&44.0$\pm$0.4&69.5$\pm$0.7&47.7$\pm$0.5&50.2$\pm$0.2&26.0$\pm$1.3&38.3\\
         && \cellcolor{gray!30}Ours&\cellcolor{gray!30}45.7$\pm$0.7 {\scriptsize \color{red}+1.7}&\cellcolor{gray!30}72.8$\pm$0.8 {\scriptsize \color{red}+3.3}&\cellcolor{gray!30}48.7$\pm$0.9 {\scriptsize \color{red}+1.0}&\cellcolor{gray!30}50.9$\pm$0.3 {\scriptsize \color{red}+0.7}&\cellcolor{gray!30}30.1$\pm$2.2{ \scriptsize \color{red}+4.1}&\cellcolor{gray!30}56.2$\pm$2.7 {\scriptsize \color{red}+17.9}\\
         \cmidrule{2-9}
         &\multirow{5}{*}{3}&FSRW \cite{metayolo} &29.2$\pm$0.4&52.7$\pm$0.6&28.5$\pm$0.4&34.2$\pm$0.3&14.2$\pm$0.7&26.8  \\
         &&FRCN-ft \cite{metarcnn}&32.1$\pm$0.5&51.3$\pm$0.8&34.3$\pm$0.6&39.1$\pm$0.5&11.1$\pm$0.9&19.0\\
         &&TFA \cite{tfa}&43.1$\pm$0.4&67.5$\pm$0.5&46.7$\pm$0.5&51.1$\pm$0.3&18.9$\pm$1.1&34.3\\
         &&DeFRCN \cite{defrcn}&45.1$\pm$0.3&70.9$\pm$0.5&48.8$\pm$0.4&50.5$\pm$0.2&29.2$\pm$1.0&52.9\\
         && \cellcolor{gray!30}Ours&\cellcolor{gray!30}46.7$\pm$0.5 {\scriptsize \color{red}+1.6}&\cellcolor{gray!30}73.7$\pm$0.5 {\scriptsize \color{red}+2.6}&\cellcolor{gray!30}50.3$\pm$0.7 {\scriptsize \color{red}+1.5}&\cellcolor{gray!30}51.4$\pm$0.4 {\scriptsize \color{red}+0.9}&\cellcolor{gray!30}32.6$\pm$1.3 {\scriptsize \color{red}+3.4}&\cellcolor{gray!30}59.7$\pm$2.1 {\scriptsize \color{red}+6.8}\\
         \cmidrule{2-9}
         &\multirow{5}{*}{5}&FSRW \cite{metayolo} &30.1$\pm$0.3&53.8$\pm$0.5&29.3$\pm$0.4&34.1$\pm$0.3&18.0$\pm$0.7&33.8  \\
         &&FRCN-ft \cite{metarcnn}&32.4$\pm$0.5&51.7$\pm$0.8&34.4$\pm$0.6&38.5$\pm$0.5&14.0$\pm$0.9&23.9\\
         &&TFA \cite{tfa}&44.1$\pm$0.3&69.1$\pm$0.4&47.8$\pm$0.4&51.3$\pm$0.2&22.8$\pm$0.9&40.8\\
         &&DeFRCN \cite{defrcn}&46.2$\pm$0.3&72.4$\pm$0.4&50.0$\pm$0.5&51.0$\pm$0.2&32.3$\pm$0.9&57.7\\
         && \cellcolor{gray!30}Ours&\cellcolor{gray!30}47.8$\pm$0.4 {\scriptsize \color{red}+1.6}&\cellcolor{gray!30}75.0$\pm$0.3 {\scriptsize \color{red}+2.6}&\cellcolor{gray!30}51.6$\pm$0.7 {\scriptsize \color{red}+1.6}&\cellcolor{gray!30}51.8$\pm$0.3 {\scriptsize \color{red}+0.8}&\cellcolor{gray!30}35.8$\pm$1.2 {\scriptsize \color{red}+3.5}&\cellcolor{gray!30}63.5$\pm$1.1 {\scriptsize \color{red}+5.8}\\
         \cmidrule{2-9}
         &\multirow{5}{*}{10}&FRCN-ft \cite{metarcnn}&33.1$\pm$0.5&53.1$\pm$0.7&35.2$\pm$0.5&38.0$\pm$0.5&18.4$\pm$0.8&44.6\\
         &&TFA \cite{tfa}&45.0$\pm$0.3&70.3$\pm$0.4&48.9$\pm$0.4&51.6$\pm$0.2&25.4$\pm$0.7&45.6\\
         &&DeFRCN \cite{defrcn}&47.0$\pm$0.3&73.3$\pm$0.3&51.0$\pm$0.4&51.3$\pm$0.2&34.7$\pm$0.7&60.8\\
         && \cellcolor{gray!30}Ours&\cellcolor{gray!30}48.5$\pm$0.3 {\scriptsize \color{red}+1.5}&\cellcolor{gray!30}75.7$\pm$0.3 {\scriptsize \color{red}+2.4}&\cellcolor{gray!30}52.5$\pm$0.6 {\scriptsize \color{red}+1.5}&\cellcolor{gray!30}52.0$\pm$0.2 {\scriptsize \color{red}+0.7}&\cellcolor{gray!30}37.9$\pm$1.1 {\scriptsize \color{red}+3.2}&\cellcolor{gray!30}66.0$\pm$1.1 {\scriptsize \color{red}+5.2}\\
         \midrule
    \end{tabular}
    }
    \end{adjustbox}

    \label{tab:gfsod}
\end{table}
\section{Qualitative Results}

\begin{figure}[hbt!]
    \centering
    \includegraphics[width=0.88\linewidth]{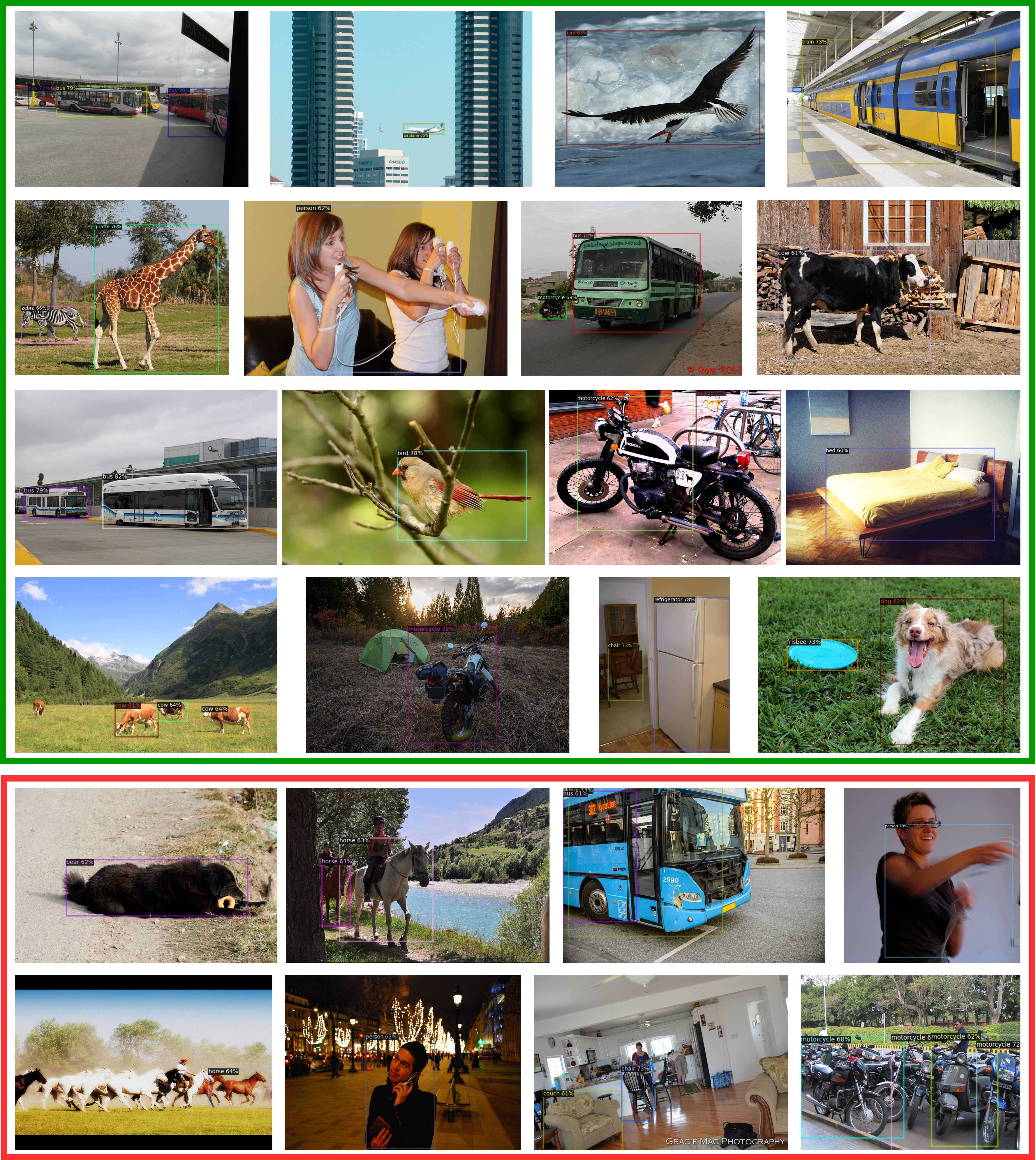}
    \caption{Qualitative result of the RISF method on the MS-COCO validation images under the gFSOD 10-shot setting. 
    We visualize bounding boxes using classification scores larger than 0.6.
    The top four rows (green box) show success scenarios while the bottom two rows (red box) show failure scenarios.
    }
    \label{fig:qualitative}
\end{figure}

Here, we show various qualitative results in Figure  \ref{fig:qualitative} on the MS-COCO 10shot datasets.
The green and red box indicates the success and failure cases of the RISF respectively.
These bounding boxes are visualized using classification scores larger than 0.6.



\section{Conclusion}

In this paper, we enhanced the prediction performance in few-shot object detection by using CLIP to measure the similarity between class names and object areas.
Furthermore, We analyzed the statistical difference in missing annotations between the clean seed and random seeds and proposed BRNL to improve performance in settings with a high prevalence of missing annotations.
Comprehensive experiments on MS-COCO and Pascal VOC demonstrate that the proposed RISF substantially outperforms the state-of-the-art approaches.
In future work, we plan to address the following limitations of our method.
1) As shown in Table 4, despite not requiring training, CM-CLIP still requires significant computational power during inference.
2) We test BNRL in specific scenarios with a high prevalence of hard negative and missing annotations. 

\bibliographystyle{unsrt}
\bibliography{risf}

\end{document}